\documentclass[letterpaper]{article} 
\usepackage{aaai23}  
\usepackage{times}  
\usepackage{helvet}  
\usepackage{courier}  
\usepackage[hyphens]{url}  
\usepackage{graphicx} 
\usepackage{natbib}  
\usepackage{caption} 
\title{Fixed-Weight Difference Target Propagation}
\author {
    Tatsukichi Shibuya, \textsuperscript{\rm 1}
    Nakamasa Inoue,      \textsuperscript{\rm 1}
    Rei Kawakami,       \textsuperscript{\rm 1}
    Ikuro Sato         \textsuperscript{\rm 1,\rm 2}
}
\affiliations {
    \textsuperscript{\rm 1} Tokyo Institute of Technology\\
    \textsuperscript{\rm 2} Denso IT Laboratory\\
    shibuya.t.ad@m.titech.ac.jp, inoue@c.titech.ac.jp, reikawa@sc.e.titech.ac.jp, isato@c.titech.ac.jp
}

\usepackage{amsmath,amssymb}
\usepackage{booktabs}
\usepackage{subcaption}
\usepackage{bm}

\usepackage[ruled,vlined]{algorithm2e}
\SetKw{KwHyper}{Hyperparameters:}
\SetKw{KwReq}{Requires:}
\SetKw{KwAll}{all}
\makeatletter
\newcommand{\removelatexerror}{\let\@latex@error\@gobble}
\makeatother

\def\ff#1{f_{#1}} 
\def\ffl#1{L_{#1}} 
\def\h#1{h_{#1}} 
\def\fb#1{g_{#1}} 
\def\tfb#1{\tilde{g}_{#1}} 
\def\fbl#1{L'_{#1}} 
\def\t#1{\tau_{#1}} 

\def\rff#1{f_{#1}} 
\def\symbolfa{\nu}
\def\symbolfb{\nu}
\def\symbolga{\mu}
\def\symbolgb{\mu}

\def\xfa#1{f_{#1}^{\symbolfa}} 
\def\xga#1{f_{#1}^{\symbolga}} 
\def\xfb#1{g_{#1}^{\symbolfb}} 
\def\xfbcheck#1{\check{g}_{#1}^{\nu}} 
\def\xgb#1{g_{#1}^{\symbolgb}} 
\def\xpa#1{\psi_{#1}} 
\def\xq#1{\mathfrak{q}_{#1}}

\def\spacexq#1{\mathfrak{Q}_{#1}}

\def\spacexfa#1{\mathcal{F}_{#1}^{\symbolfa}}
\def\spacexfb#1{\mathcal{G}_{#1}^{\symbolfb}}
\def\spacexga#1{\mathcal{F}_{#1}^{\symbolga}}
\def\spacexgb#1{\mathcal{G}_{#1}^{\symbolgb}}

\newcommand{\argmin}{\mathop{\rm argmin}\limits}

\def\rfb#1{g_{#1}} 

\def\spacexfbr#1{\mathfrak{G}_{#1}}
\def\mathsize{}

\begin{document}
\maketitle
\begin{abstract}
Target Propagation (TP) is a biologically more plausible algorithm than the error backpropagation (BP) to train deep networks, and improving practicality of TP is an open issue. TP methods  require the feedforward and feedback networks to form layer-wise autoencoders for propagating the target values generated at the output layer. However, this causes certain drawbacks; {\em e.g.}, careful hyperparameter tuning is required to synchronize the feedforward and feedback training, and frequent updates of the feedback path are usually required than that of the feedforward path. Learning of the feedforward and feedback networks is sufficient to make TP methods capable of training, but is having these layer-wise autoencoders a necessary condition for TP to work? We answer this question by presenting {\it Fixed-Weight Difference Target Propagation} (FW-DTP) that keeps the feedback weights constant during training. We confirmed that this simple method, which naturally resolves the abovementioned problems of TP, can still deliver informative target values to hidden layers for a given task; indeed, FW-DTP consistently achieves higher test performance than a baseline, the Difference Target Propagation (DTP), on four classification datasets. We also present a novel propagation architecture that explains the exact form of the feedback function of DTP to analyze FW-DTP.
\end{abstract}
\section{Introduction}
Artificial Neural Networks (NNs) were introduced to model the information processing in the neural circuits of the brain \cite{fitch_1944,rosenblatt58}.
The error backpropagation (BP) has been the most widely used algorithm to optimize parameters of multi-layer NNs with gradient descent \cite{bp}, but the lack of consistency with neuroscientific findings has been pointed out \cite{cric,glorot}. 
In particular, the inconsistencies include that 
1) in BP, the feedback path is the reversal of the feedforward path in a way that the same synaptic weight parameters are used (a.k.a. weight transport problem \cite{transport}), while the brain most likely uses different sets of parameters in the feedforward and feedback processes; 
2) in BP, the layer-to-layer operations are asymmetric between feedforward and feedback processes ({\em i.e.}, the feedback process does not require the activation used in the feedforward process), while the brain requires symmetric operations.
Although there are on-going research efforts to connect the brain and BP \cite{brain}, many researchers seek less inconsistent yet practical algorithms of network training \cite{fa,tp,dtp,dttp,theoretical,gait,equilibrium} because biologically plausible algorithms that may bridge the gap between neuroscience and computer science are believed to enhance machine learning.

Feedback alignment (FA) \cite{fa} was proposed to resolve the weight transport problem by using fixed random weights for error propagation.
It is worth noting that FA has been shown to outperform BP on real datasets, although the results are somewhat outdated \cite{dfa, dfa-new}.

Target propagation (TP) \cite{tp, tp-lecun} has been proposed as a NN training algorithm that can circumvent the inconsistencies 1) and 2).
The main idea of TP is to define target values for hidden neurons in each layer in a way that the \textit{target values} (not the \textit{error}) are backpropagated from the output layer down to the first hidden layer, using the same activation function used in the feedforward process.
The feedback network, which does \textit{not} share parameters with the feedforward network, is trained so that each layer becomes an approximated inverse of the corresponding layer of the feedforward network, whereas the parameters of the feedforward network are updated to achieve the layer-wise target.
In TP, the feedback network ideally realizes layer-wise autoencoders with the feedforward network, but in reality it often ends up with imperfect autoencoders, which could cause optimization problems \cite{dtp, theoretical}.
Among the methods that alleviate such small discrepancies \cite{dtp, dttp}, difference target propagation (DTP) \cite{dtp} introduces linear correction terms to the feedback process and significantly improved the recognition performance of TP.

However, while the formalism of DTP to compute layer-wise targets with a feedback network is theoretically sound, training the feedback network is often demanding in the following aspects:
a) Synchronous training of the feedforward and feedback networks often requires careful hyperparameter tuning \cite{assessing}.
b) Training of the feedback network could be computationally very expensive. 
According to previous work \cite{assessing, theoretical, scaling}, weight updates of the feedback networks were more frequent than those of the feedforward networks.
In the latest research \cite{scaling}, the number of updating feedback weights is set to several tens of times of that of the feedforward weights. 
For these reasons, training a feedback network typically requires a large cost including hyperparameter tuning. 

It is clear that having a relation of layer-wise autoencoders by the feedforward and feedback network is sufficient for the target propagation algorithms to gain training capability.
In this work, we aim to answer the question whether constructing layer-wise autoencoders is also a \textit{necessary} condition for target propagation to work.
To answer this question, we examined a very simple approach, where the parameters of the feedback network are kept fixed while the feedforward network is trained just as DTP.
No reconstruction loss is imposed, so the feedforward and feedback networks are not forced to form autoencoders.
Nevertheless, our new target propagation method, {\it fixed-weight difference target propagation} (FW-DTP), greatly improves the stability of training and test performance compared to DTP while reducing computational complexity from DTP.
The idea of fixing feedback weights is inspired by FA, which fixes feedback weights during BP to avoid the weight transport problem.
But the difference is that FW-DTP greatly simplifies the learning rule of DTP by removing layer-wise autoencoding losses, whereas FA has no such effect.
We provide mathematical expressions about conditions that network trained with DTP will implicitly acquire with and without fixing feedback weights.
We further propose a novel propagation architecture which can explicitly provide the exact form of the feedback function of DTP, which implies that FW-DTP acquires implicit autoencoders.
It is worth mentioning that Local Representation Alignment (LRA) \cite{lra} (followed by \cite{dtpsigma, lra2}) also proposed biologically-plausible layer-wise learning rules with fixed parameters, though it does not belong to the target propagation family.

Our contribution is three-fold:
\textbf{1)} We propose {\it fixed-weight difference target propagation} (FW-DTP) that fixes feedback weights and drops the layer-wise autoencoding losses from DTP. 
Good learnability of FW-DTP indicates that optimizing the objectives of layer-wise target reconstruction is not necessary for the concept of target propagation to properly work.
\textbf{2)} We present a novel architecture that explicitly shows the exact form of feedback function of DTP, which allows for an accurate notation of how the targets are backpropagated in the feedback network.
\textbf{3)} We experimentally show that FW-DTP not only improves the stability of training from DTP, but also improves the mean test accuracies from DTP on four image classification datasets just like FA outperforming BP by using fixed backward weights.
\section{Overview: Target Propagation Methods}\label{seq:bg}
We overview target propagation methods including TP \cite{tp} and DTP \cite{dtp}.

\noindent {\bf Definition \ref{seq:bg}.1 (Feedforward and feedback functions).}
Let $\mathcal{X}$ and $\mathcal{Y}$ be the input and output spaces, respectively.
A feedforward function $F: \mathcal{X} \to \mathcal{Y}$ is defined as a composite function of layered encoders
$\ff{l}~(l=1,\cdots L)$
by
\mathsize\begin{align}\label{eq:forward-def}
F(x) = \ff{L} \circ \ff{L-1} \circ \cdots \circ \ff{1}(x)
\end{align}\normalsize
where $L$ is the number of encoding layers, $x \in \mathcal{X}$ is an input.
A feedback function $G: \mathcal{Y} \to \mathcal{X}$ is defined by
\mathsize\begin{align}\label{eq:backward-def}
G(y) = \fb{1} \circ \fb{2} \circ \cdots \circ \fb{L} (y)
\end{align}\normalsize
where $y \in \mathcal{Y}$ is an output and $\fb{l}$ is the $l$-th decoder.
Each $\fb{l}$ is paired with $\ff{l}$, and will be trained to approximately invert $\ff{l}$.
The feedforward activation $\h{l}$ is recursively defined as
\mathsize\begin{align}\label{eq:layer-forward-def}
\h{l} = 
\begin{cases}
x & (l = 0)\\
\ff{l}(\h{l-1}) & (l = 1, \cdots, L)
\end{cases}
\end{align}\normalsize
and the target $\t{l}$ is recursively defined in the descending order as
\mathsize\begin{align}
\label{eq:layer-backward-def}
\t{l} = 
\begin{cases}
y^{\star} & (l = L)\\
\tfb{l+1}(\t{l+1}) & (l = L-1, \cdots, 0)
\end{cases}
\end{align}\normalsize
where $y^{\star}$ is the output target. In Eq.~(\ref{eq:layer-backward-def}), $\tfb{l}$ is an extended function of $\fb{l}$ to propagate targets, and it could be the same as $\fb{l}$.
Note that this paper focuses on supervised learning where loss $\mathcal{L}(F(x), y)$ to be minimized takes finite values over all training pairs $(x, y) \in \mathcal{X} \times \mathcal{Y}$.

\noindent {\bf Target Propagation (TP).} TP is an algorithm to learn the feedforward and feedback function where $\ff{l}$ and $\fb{l}$ are parameterized. It defines the output target based on gradient descent (GD) \cite{tp} as
\mathsize\begin{align}
y^{\star}(h_L) = \h{L} - \beta \frac{\partial \mathcal{L}(\h{L}, y)}{\partial \h{L}}
\end{align}\normalsize
where $\beta$ is a nudging parameter.
For propagating targets, $\tfb{l} = \fb{l}$ is used.
TP updates feedforward weights (the parameters of $f_{l}$) and feedback weights (the parameters of $g_{l}$) alternately.
The $l$-th layer's feedforward weight is updated to reduce layer-wise local loss:
\mathsize\begin{align}
\label{eq6}
\ffl{l} = \frac{1}{2 \beta} \| \h{l} - \t{l} \|_{2}^{2}
\end{align}\normalsize
where $\tau_l$ is considered as a constant with respect to the $l$-th layer's feedforward weight,
{\em i.e.}, the gradient of $\t{l}$ with respect to the weight is $0$.
The $l$-th layer's feedback weight is updated to reduce reconstruction loss:
\mathsize\begin{align}
\label{eq:rec}
\fbl{l} = \frac{1}{2}~\mathbb{E}_{\epsilon \sim \mathcal{N}(0, \sigma^{2}I)} \left[ \| \h{l-1} + \epsilon - \fb{l} \circ \ff{l} (\h{l-1} + \epsilon) \|_{2}^{2}\right]
\end{align}\normalsize
where $\epsilon$ is a small noise to improve the robustness of inversion. A known limitation of TP 
is that imperfectness of the feedback function as inverse leads to a critical optimization problem \cite{dtp,theoretical}, {\it i.e.}, the update direction $\tau_l-h_l$ involves reconstruction errors $g_{l+1}(f_{l+1}(h_{l}))-h_{l}$; thus, the feedforward network is not trained properly with an imprecisely optimized feedback network.

\noindent {\bf Difference Target Propagation (DTP).}
Lee {\em et al.} \cite{dtp} show that {\it difference correction}, subtracting the difference $g_{l+1}(\h{l+1}) - \h{l}$ from the target, alleviates the limitation of TP, and they introduce DTP, whose function $\tfb{l+1}$ for propagating targets in Eq.~(\ref{eq:layer-backward-def}) is defined by
\mathsize\begin{align}\label{eq:dtp-propagation}
\tfb{l+1}(\t{l+1}) = \fb{l+1}(\t{l+1}) + \h{l} - g_{l+1}(\h{l+1}).
\end{align}\normalsize
The losses for updating feedforward and feedback weights are the same as those of TP. In DTP, assuming all encoders are invertible, the first order approximation of $\Delta h_l := \tau_l-h_l$ is given by
\mathsize\begin{align}\label{eq:dtp-update}
\Delta h_l
& \simeq \left[\prod_{k=l+1}^{L} \frac{\partial g_k(h_k)}{\partial h_k} \right]\Delta h_L \\
&= \left[\prod_{k=l+1}^{L} J_{g_k} \right] \left( -\beta \frac{\partial \mathcal{L}(h_L, y)}{\partial h_L}\right)\\
&= -\beta J_{f_{l+1:L}}^{-1}\frac{\partial \mathcal{L}(h_L, y)}{\partial h_L}
\end{align}\normalsize
where $J_{g_k}:=\partial g_k(h_k)/\partial h_k$ is the Jacobian matrix of $g_k$ evaluated at $h_k$. Here, $J_{g_l} = J_{f_l}^{-1}~(l=1, \cdots ,L)$ and $J_{f_{l+1:L}}^{-1} = \prod_{k=l+1}^{L} J_{f_k}^{-1}$ are used due to the invertibility, where $J_{f_l}:=\partial f_k(h_{k-1})/\partial h_{k-1}$ is the Jacobian matrix of $f_k$ evaluated at $h_{k-1}$.
The notation $()_{a:b}$ is for 
composing functions from layers $a$ to $b$, \;{\em e.g.},\; $f_{l+1:L} = f_L\circ\dots\circ f_{l+1}$.
The update rule of DTP is regarded as a hybrid of GD and Gauss-Newton (GN) algorithm \cite{gn}.
Note that, in the case of the {\it non-invertible} encoders, DTP obtains
the condition $J_{g_l}=J_{f_l}^{+}$ where $J_{f_{l}}^{+}$ is the Moore-Penrose inverse \cite{moore, penrose} of $J_{f_l}$, however, $J_{f_{l+1:L}}^{+} = \prod_{k=l+1}^{L} J_{f_k}^{+}$ is not always satisfied \cite{theoretical, factrize}.
\section{Proposed Method}
\label{seq:FW-DTP}
This section presents the proposed fixed-weight difference target propagation (FW-DTP) that drops the training of feedback weights.
We first propose FW-DTP according to the traditional notation (defined in Section 2) in \ref{sec4.1}.
We then analyze FW-DTP from two points of view:
the conditions for Jacobians in \ref{sec4.4} and
the exact form of the feedback function in \ref{sec4.2}.
From these analyses,
we explain why fixed-weights of FW-DTP has a good learnability. 

\subsection{Fixed-Weight Difference Target Propagation}\label{sec4.1}
FW-DTP is defined as the algorithm that omits reconstruction loss for updating feedback weights in DTP. All feedback weights are first randomly initialized
and then fixed during training.
For example,
with a fully connected network,
the $l$-th encoder and decoder of FW-DTP are defined by
\mathsize\begin{align}
    f_l(h_{l-1}) := \sigma_l(W_lh_{l-1}),~~
    g_l(\tau_l) := \sigma_l'(B_l\tau_l)
\end{align}\normalsize
where $\sigma$ and $\sigma'$ are non-linear activation functions and $W_l$ and $B_l$ are matrices which denote the feedforward and feedback weights, respectively.
$B_l$ is first initialized with a distribution $P(B_l)$ and then fixed, while $W_l$ is updated in the learning process. The feedback propagation of targets are defined by Eq.~(\ref{eq:dtp-propagation}). 
Note that DTP asymptotically approaches FW-DTP by decreasing the learning rate of the feedback weights. This is discussed with experimental results in the appendix.

\subsection{Analysis 1: Condition for Jacobians}\label{sec4.4}
Here, we discuss conditions for DTP to appropriately work. 
Given that precise inverse relation between $f_l$ and $g_l$ may not be always obtainable in DTP, 
training with inaccurate targets can degrade the overall performance of the feedforward function. 
Now, consider two directions $\tau_{l}-h_{l}$ and $f_{l}(h^{*}_{l-1}) - h_{l}$,
a vector from the activation $h_{l}$ to the target $\tau_{l}$ at layer $l$, and another from $h_{l}$ to the point $f_{l}(h^{*}_{l-1})$. If the condition 
\mathsize\begin{align}\label{eq:angle}
     -\frac{\pi}{2} \le \angle (\tau_{l}-h_{l}, f_{l}(h^{*}_{l-1})-h_{l}) \le \frac{\pi}{2}
     \text{~~where~~} h^{*}_{l-1} = \tau_{l-1}
\end{align}\normalsize
holds, {\em i.e.}, if the angle between them is within 90 degrees,
the loss of this sample is expected to decrease
because $f_{l}(h^{*}_{l-1})$ is the best point achieved by learning $(l-1)$-th encoder.
By applying the first order approximation, Eq.~(\ref{eq:angle}) is rewritten as
\mathsize\begin{align}\label{eq:strict-condition}
    \Delta \h{l}^{\top} J_{f_{l}}J_{g_{l}} \Delta \h{l} \ge 0
\end{align}\normalsize
therefore, the sufficient condition of Eq.~(\ref{eq:angle}) is that $J_{f_{l}}J_{g_{l}}$ is a positive semi-definite matrix. As Table~\ref{tab:compare-condition} shows, minimization of reconstruction losses of DTP such as original DTP (Eq.~(\ref{eq:rec})), difference reconstruction loss (DRL) \cite{theoretical} and local difference reconstruction loss (L-DRL) \cite{scaling} naturally satisfy the positive semi-definiteness by enforcing the Jacobian matrix $J_{g_{l}}$ as the inverse or transpose of $J_{f_{l}}$.
On the other hand, positive semi-definiteness requires
\mathsize\begin{align}\label{eq:min-condition}
    \inf_{\epsilon}\left[\epsilon^{\top} J_{f_{l}}J_{g_{l}} \epsilon\right]  \ge 0
\end{align}\normalsize
however, this condition could be somewhat too strict, given that features may not always span the full space.
In FW-DTP, the strict condition expressed in Eq.~(\ref{eq:min-condition}) is not generally satisfied because FW-DTP has no feedback objective function to learn to explicitly satisfy this condition.
Now, let us consider a hypothetical situation where the product of Jacobians satisfies the condition,
\mathsize\begin{align}\label{eq:probabilistic-condition}
    \mathbb{E}_{\epsilon \sim p(\cdot)}\left[\epsilon^{\top} J_{f_{l}}J_{g_{l}} \epsilon\right]  \ge 0
\end{align}\normalsize
where the infimum in Eq.~(\ref{eq:min-condition}) is replaced with the expectation over some origin-centric rotationally-symmetric distribution $p(\cdot)$ such as a zero-mean isotropic Gaussian distribution.
Then, it is straightforward to show that Eq.~(\ref{eq:probabilistic-condition}) is equivalent to
\mathsize\begin{align}\label{eq:trace}
    \mathrm{tr}(J_{f_{l}}J_{g_{l}}) \ge 0
\end{align}\normalsize
(the proof is in Appendix \ref{app:proof}).
The condition expressed in Eq.~(\ref{eq:trace}) is weaker than Eq.~(\ref{eq:min-condition}).
Under the condition of Eq.~(\ref{eq:trace}), if $(l-1)$-th activation moves toward the target, 
it will shifts $l$-th activation toward the corresponding target within $\pi/2$ range as the expectation (over $p$).
Although the condition of Eq.~(\ref{eq:trace}) is somewhat artificial, but indeed we found that FW-DTP does satisfy this condition in our experiment as we show in Section 4.
The condition expressed in Eq.~(\ref{eq:trace}) could be regarded as a type of alignments that the network implicitly acquires when its feedback weights are fixed during DTP updates.

\begin{table*}[t]
    \caption{The conditions of the Jacobians obtained by various reconstruction losses and FW-DTP.}
    \label{tab:compare-condition}
    \begin{center}\begin{small}\begin{sc}
    \begin{tabular}{lcccc}
        \toprule
        Method & DTP 
        &DRL 
        &L-DRL 
        &FW-DTP\\
        \midrule
        Condition
        &$J_{g_{l}} = J_{f_l}^{+}$
        &$\prod_{k=l}^{L} J_{g_k} = J_{f_{l:L}}^{+}$
        &$J_{g_{l}} = J_{f_l}^{\top}$
        &$\mathrm{tr}(J_{f_l}J_{g_{l}})>0$\\
        \bottomrule
    \end{tabular}
    \end{sc}\end{small}\end{center}
\end{table*}

\subsection{Analysis 2: Exact Form of Feedback Function
}
\label{sec4.2}\label{sec4.3}
To show how targets are propagated in FW-DTP,
we present a propagation architecture
which 
provides the exact form of the feedback function of DTP.
There exists
no autoencoders in FW-DTP at least explicitly;
however, difference correction creates autoencoders implicitly. 
To explicitly show this, instead of using the function $\tilde{g}_{l}$ for propagating targets in Eq~(\ref{eq:layer-backward-def}), we decomposed encoder and decoder as $f_{l} = f^{\nu}_{l} \circ f^{\mu}_{l}$ and $g_{l} = g^{\nu}_{l} \circ g^{\mu}_{l}$
to incorporate the difference correction mechanism into $g^{\nu}_{l}$.
Using the proposed architecture represented by Eqs.~(\ref{eq:layer-backward-def-re}-\ref{eq:shortcut-function}), TP and DTP are reformulated as Eqs.~(\ref{eq:tp-formulation}-\ref{eq:objectivedtp}) and the training process is also reformulated as Eq.~(\ref{eq:objective}).

\begin{figure*}[t]
    \centering
    \includegraphics[width=16.0cm]{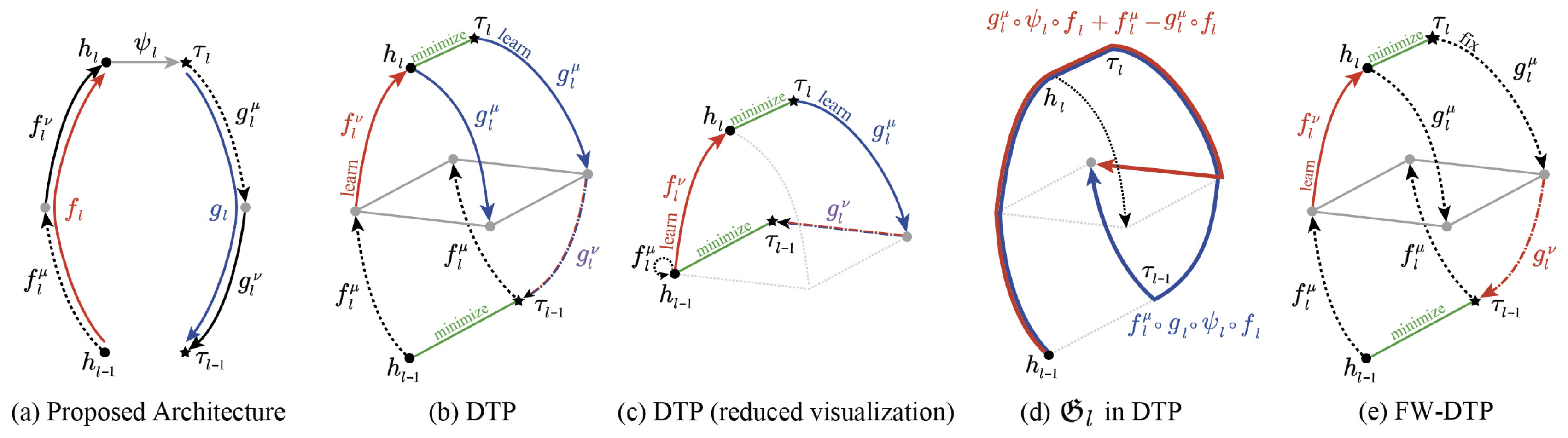}
    \caption{Proposed propagation architecture and its reduction to DTP. (a) The proposed architecture. The encoder $f_{l}$ is decomposed into $f^{\mu}_{l}$ and $f^{\nu}_{l}$. The decoder $g_{l}$ is decomposed into $g^{\mu}_{l}$ and $g^{\nu}_{l}$. $\psi_{l}$ is the shortcut function from an activation $h_{l}$ to the target $\tau_{l}$. (b) Reduction to DTP. $g^{\nu}_{l}$ is a function of difference correction. $f^{\mu}_{l}$ is illustrated as non-identity function. (c) Reduction to DTP, where $f^{\mu}_{l}$ is illustrated as the identity function. This is the well-known visualization of DTP. (d) The search space $\mathfrak{G}_{l}$ for $g^{\nu}_{l}$. (e) FW-DTP with fixed $g^{\mu}_{l}$}.
    \label{figB}
\end{figure*}

\noindent {\bf Definition \ref{seq:FW-DTP}.1 (Propagation Architecture).}
We define a feedforward function $F: \mathcal{X} \to \mathcal{Y}$ with encoders $\rff{l}$ and a feedback function $G: \mathcal{Y} \to \mathcal{X}$ with decoders $\rfb{l}$ by Eqs.~(\ref{eq:forward-def}-\ref{eq:backward-def}). The targets are recursively defined in the descending order as
\mathsize\begin{align}
\label{eq:layer-backward-def-re}
\t{l} = 
\begin{cases}
y^{\star} & (l = L)\\
\rfb{l+1}(\t{l+1}) & (l = L-1, \cdots, 0)
\end{cases}
\end{align}\normalsize
where $y^{\star}$ is the output target. Eq~(\ref{eq:layer-backward-def-re}) differs from Eq~(\ref{eq:layer-backward-def}) in that we avoid to define $\tilde{g}_{l}$. 
Further, we introduce four functions
$\xga{l}, \xfa{l}, \xgb{l},\xfb{l}$
that decompose the encoder and decoder into
\mathsize\begin{align}
\rff{l} = \xfa{l} \circ \xga{l},~~
\rfb{l} = \xfb{l} \circ \xgb{l}.
\end{align}\normalsize
We also define a shortcut function $\xpa{l}$ that map the activation to the target as
\mathsize\begin{align}\label{eq:shortcut-function}
\xpa{l} (\h{l}) =
\begin{cases}
\t{L} & (l = L)\\
\rfb{L:l+1}\circ\xpa{L}\circ\rff{l+1:L}(\h{l}) & (l = L-1, \cdots, 0).
\end{cases}
\end{align}\normalsize
Here, $\xpa{l}(\h{l})=\t{l}$. Figure~\ref{figB}a illustrates the proposed propagation architecture.
With this architecture, we expect that $g_{l} \circ f_{l}$ will become an autoencoder after convergence with the activations sufficiently close to the corresponding targets .
It is reduced to DTP when
$f^{\mu}_{l}$ is the identity function,
$f^{\nu}_{l}$ is a parameterized function ({\it e.g.}, $f^{\nu}_{l}(h_{l-1}) = \sigma(W_{l} h_{l-1})$),
$g^{\mu}_{l}$ is another parameterized function,
and $g^{\nu}_{l}$ is a function of difference correction,
as shown in Figure~\ref{figB}b~and~\ref{figB}c. Note that Figure~\ref{figB}c is a well-known visualization of DTP \cite{dtp}.
The main problem we would like to discuss is whether there exists the exact form of $g^{\nu}_{l}$.
With the traditional notations in Eq.~(\ref{eq:dtp-propagation}), $\tilde{g}_{l+1}$ is defined as a function of $\tau_{l+1}$, however, it uses $h_{l}$ and $h_{l+1}$ in the right side of the equation.
This makes it difficult to analyze the shape of feedback function; thus, we define the training process here as follows.

\noindent {\bf Definition \ref{seq:FW-DTP}.3 (Training).}
Let $\xq{l} = (\xga{l}, \xfa{l}, \xgb{l},  \xfb{l})$ a quadruplet of functions.
We define training as the process to solve the following layer-wise problem:
\mathsize\begin{align}
\label{eq:objective}
\xq{l}^{*} = \argmin_{\xq{l} \in \spacexq{l}} \mathcal{O}_l
\end{align}\normalsize
where $\spacexq{l} = \spacexga{l} \times \spacexfa{l} \times \spacexgb{l} \times \spacexfb{l}$ is a function space (search space), and $\mathcal{O}_l$ is the objective function.

This definition
involves TP and DTP variants as follows.
 
\noindent {\bf Target Propagation.}
Using the proposed architecture, TP is defined as a training process with the 
search spaces:
\mathsize\begin{align}\label{eq:tp-formulation}
\spacexga{l} = \{ \textit{id} \},~
\spacexfa{l} = \{ p_{\theta} : \theta \in \Theta_{l} \}\\
\spacexgb{l} = \{ p_{\omega} : \omega \in \Omega_{l} \},~
\spacexfb{l} = \{ \textit{id} \}
\end{align}\normalsize
where $\textit{id}$ is the identity function and $p_{\theta}$ and $p_{\omega}$ are parameterized functions with learnable parameters $\theta$ and $\omega$, respectively.
$\Theta_{l}$ and $\Omega_{l}$ are the parameter spaces.
TP solves Eq.~(\ref{eq:objective}) by alternately solving two problems:
\mathsize\begin{align}
\label{eq:objectivetp}
f^{\nu*}_{l} = \argmin_{f^{\nu}_{l} \in \mathcal{F}^{\nu}_{l}} \mathcal{O}^{(1)}_{l}\\
g^{\mu*}_{l} = \argmin_{g^{\mu}_{l} \in \mathcal{G}^{\mu}_{l}} \mathcal{O}^{(2)}_{l}
\end{align}\normalsize
where $\mathcal{O}^{(1)}_{l}$ is the layer-wise local loss in Eq.~(\ref{eq6}) and $\mathcal{O}^{(2)}_{l}$ is the reconstruction loss in Eq.~(\ref{eq:rec}).

\noindent {\bf Difference Target Propagation.}
DTP is also defined with a search space $\spacexfbr{l}$ for $g^{\nu}_{l}$ as follows:
\mathsize\begin{align}
\spacexga{l} = \{ \textit{id} \},~
\spacexfa{l} = \{ p_{\theta} : \theta \in \Theta_{l} \}\\
\spacexgb{l} = \{ p_{\omega} : \omega \in \Omega_{l} \},~
\spacexfb{l} = \spacexfbr{l}
\end{align}\normalsize
\mathsize\begin{align}
\label{eq:setfb}
\text{where~~~}
\spacexfbr{l}
= \{&\xfb{l}: 
d_{P}(\xga{l} \circ \rfb{l} \circ \xpa{l} \circ \rff{l}, \nonumber\\
&\xgb{l} \circ \xpa{l} \circ \rff{l} + \xga{l} - \xgb{l} \circ \rff{l}) = 0\}
\end{align}\normalsize
and $d_{P}$ with norm $P$ ({\em e.g.}, $L_{2}$ norm) is a distance in the function space.

Figure~\ref{figB}d shows the two functions $\xga{l} \circ \rfb{l} \circ \xpa{l} \circ \rff{l}$ and $\xgb{l} \circ \xpa{l} \circ \rff{l} + \xga{l} - \xgb{l} \circ \rff{l}$ in blue and red, respectively; namely,  $\spacexfbr{l}$ is the function subspace of $\xfb{l}$ where these two functions (the blue and red arrows in \ref{figB}d) are equal.
By assuming functions $\xfa{l}, \xpa{l}, \xgb{l}$ are bijective, we have $ \spacexfbr{l} = \{ \xfbcheck{l} \}$ where
\mathsize\begin{align}
\label{eq:invfb}
\xfbcheck{l} = 
\textit{id} 
&+ (\xfa{l})^{-1} \circ (\xpa{l})^{-1} \circ (\xgb{l})^{-1}\nonumber \\
&- \xgb{l} \circ (\xpa{l})^{-1} \circ (\xgb{l})^{-1}.
\end{align}\normalsize
This is the exact form of difference correction in our formulation.
This shows that $g^{\nu}_{l}$ is implicitly updated by updating $f^{\nu}_{l}$ and $g^{\mu}_{l}$.
Therefore, DTP solves Eq.~(\ref{eq:objective}) by alternately solving two problems:
\mathsize\begin{align}
\label{eq:objectivedtp}
(f^{\nu*}_{l}, g^{\nu*}_{l}) = \hspace{-9pt} \argmin_{(f^{\nu}_{l}, g^{\nu}_{l}) \in \mathcal{F}^{\nu}_{l} \times \mathcal{G}^{\nu}_{l}}
\hspace{-7pt}
\mathcal{O}^{(1)}_{l},~
(g^{\mu*}_{l}, g^{\nu*}_{l}) = \hspace{-9pt}
\argmin_{(g^{\mu}_{l},g^{\nu}_{l}) \in \mathcal{G}^{\mu}_{l} \times \mathcal{G}^{\nu}_{l}}
\hspace{-7pt}
\mathcal{O}^{(2)}_{l}
\end{align}\normalsize
where the objective function is the same as that of TP.
Eq.~(\ref{eq:objectivedtp}) indicates that updating the feedforward weights implicitly update $g^{\nu}_{l}$ in the feedback path.

\noindent {\bf Fixed-Weight Difference Target Propagation.}
From Eq.~(\ref{eq:invfb}), we notice that {\it DTP works even with fixed} $\xgb{l}$ because $\xfb{l}$ is updated in conjunction with $\xfa{l}$.
If the function space $\mathcal{F}^{\nu}_{l}$ is large enough for finding an appropriate pair of $f^{\nu}_{l}$ and $g^{\nu}_{l}$,
parametrization of the two function spaces $\spacexfa{l}$ and $\spacexgb{l}$ may be redundant.
Based on this observation,
FW-DTP uses a unit set for $\spacexgb{l}$:
\mathsize\begin{align}
\label{eqp}
\spacexga{l} = \{ \textit{id} \},~
\spacexfa{l} = \{ p_{\theta} : \theta \in \Theta_{l} \},~
\spacexgb{l} = \{ r_{l} \},~
\spacexfb{l} = \spacexfbr{l}
\end{align}\normalsize
where $r_{l}$ is a fixed random function. FW-DTP solves Eq.~(\ref{eq:objective}) by  solving one problem:
\mathsize\begin{align}
\label{eq:objectivefwdtp}
(f^{\nu*}_{l}, g^{\nu*}_{l}) = \hspace{-9pt} \argmin_{(f^{\nu}_{l}, g^{\nu}_{l}) \in \mathcal{F}^{\nu}_{l} \times \mathcal{G}^{\nu}_{l}}
\hspace{-7pt}
\mathcal{O}^{(1)}_{l}.
\end{align}\normalsize
Figure~\ref{figB}e shows that in FW-DTP, $g^{\mu}_{l}$ is fixed but $g^{\nu}_{l}$ colored in red moves with $f^{\nu}_{l}$, and thus there still exists an autoencoder $g_{l} \circ f_{l}$. This is one of the reasons why FW-DTP has an ability to propagate targets to decrease loss.
To keep non-linearity and the ability to entangle elements from different dimension on the feedback path, $r_{l}(a) = \sigma(B_{l}a)$ would be the simplest choice where $B_{l}$ is a random matrix fixed before training and $\sigma$ is a non-linear activation function.
FW-DTP is more efficient than DTP because it reduces the number of learnable parameters.
\section{Experiments} 
In this section, we show experimental results.
\footnote{Our code is available at \url{https://github.com/TatsukichiShibuya/Fixed-Weight-Difference-Target-Propagation}.}
First, we show that the weak condition expressed in Eq.~(\ref{eq:probabilistic-condition}) is satisfied by FW-DTP experimentally. We then compare FW-DTP with TP and DTP variants. Lastly, we evaluate the hyperparameter sensitivity and computational cost, and show that FW-DTP is more stable and computationally efficient than DTP.

\subsection{Weak and Strict Conditions of Jacobians}\label{ex1}
\noindent {\bf Experimental set-up.}
This experiment aims to show that FW-DTP satisfies
the weak condition of Jacobians given by Eq.~(\ref{eq:probabilistic-condition}) during its training process.
We also show that FW-DTP does not satisfy the strict condition expressed in Eq.~(\ref{eq:min-condition}) in contrast to DTP.

Evaluation details are as follows.
For the weak condition,
we
directly measured the trace of $J_{f_{l}}J_{g_{l}}$ (With notations in Analysis 2, this is  $J_{f^{\nu}_{l}}J_{g^{\mu}_{l}}$).
For the strict condition, we measured the proportion of the number of non-negative eigenvalues of $J_{f_{l}}J_{g_{l}}$ to its dimension. 
This is a measure of positive semi-definiteness.
The MNIST dataset \cite{mnist} was used for this evaluation.
A fully connected network with 6 layers each with 256 units was trained with cross-entropy loss. Note that the first and the last encoders are non-invertible due to the difference of the input and output dimensions. We chose the hyperbolic tangent as the activation function, but only for FW-DTP, batch normalization (BN) \cite{bn} was applied after each hyperbolic tangent. The necessity and effectiveness of BN is discussed in the appendix. 
Stochastic gradient descent (SGD) was used as the optimizer. Details of the hyperparameters are also provided in the appendix.
The feedforward and feedback weights were initialized with random orthogonal matrices and random numbers from uniform distribution $U(-0.01, 0.01)$, respectively.

\begin{figure*}[t]
    \hspace{-0.7em}
    \begin{tabular}{cc}
      \begin{minipage}{0.5\hsize}
        \includegraphics[keepaspectratio, scale=0.25]{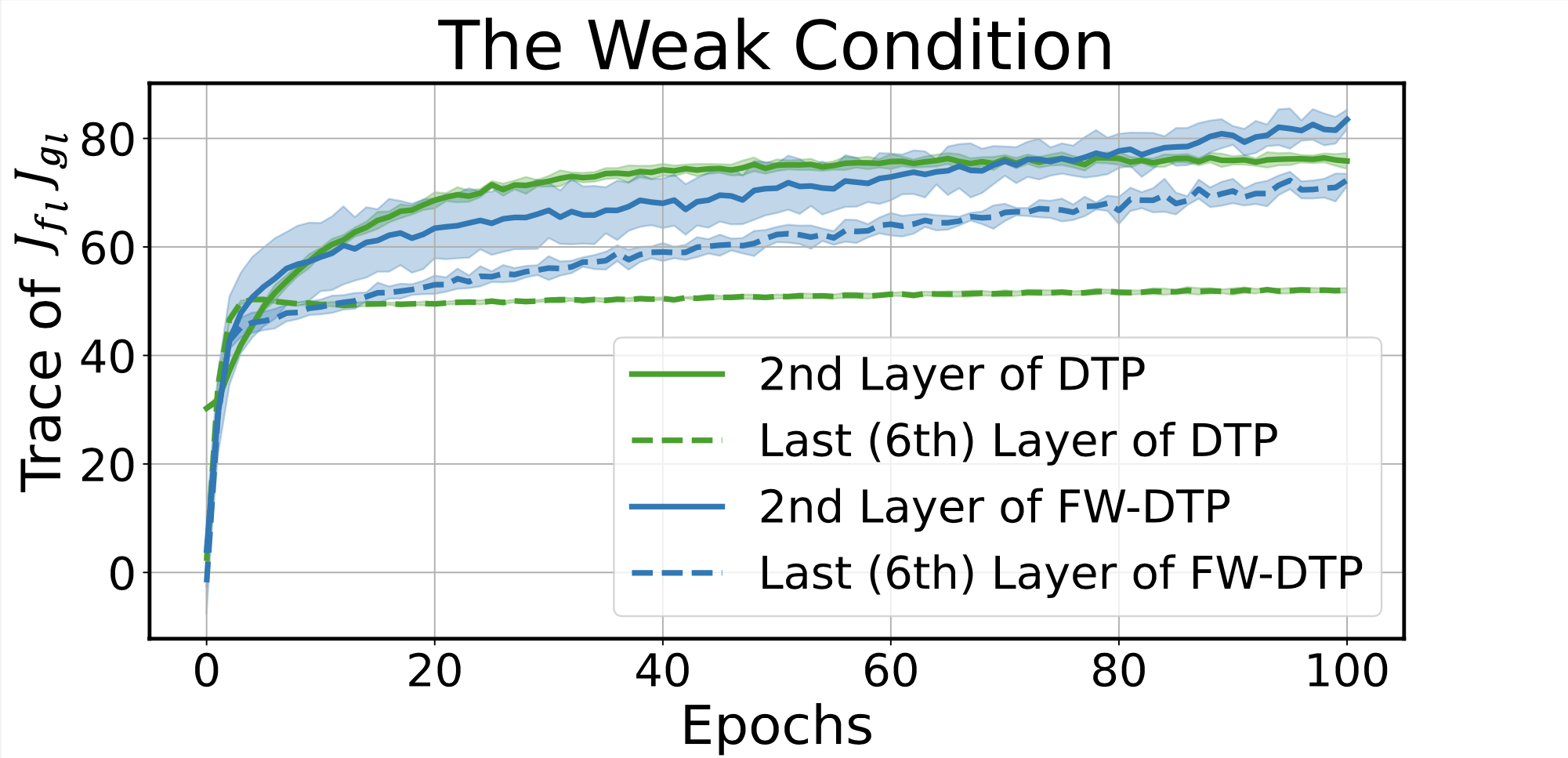}
        \subcaption{Trace of $J_{f_{l}}J_{g_{l}}$}
        \label{fig:probabilistic-condition}
      \end{minipage}
      &
      \begin{minipage}{0.5\hsize}
        \includegraphics[keepaspectratio, scale=0.25]{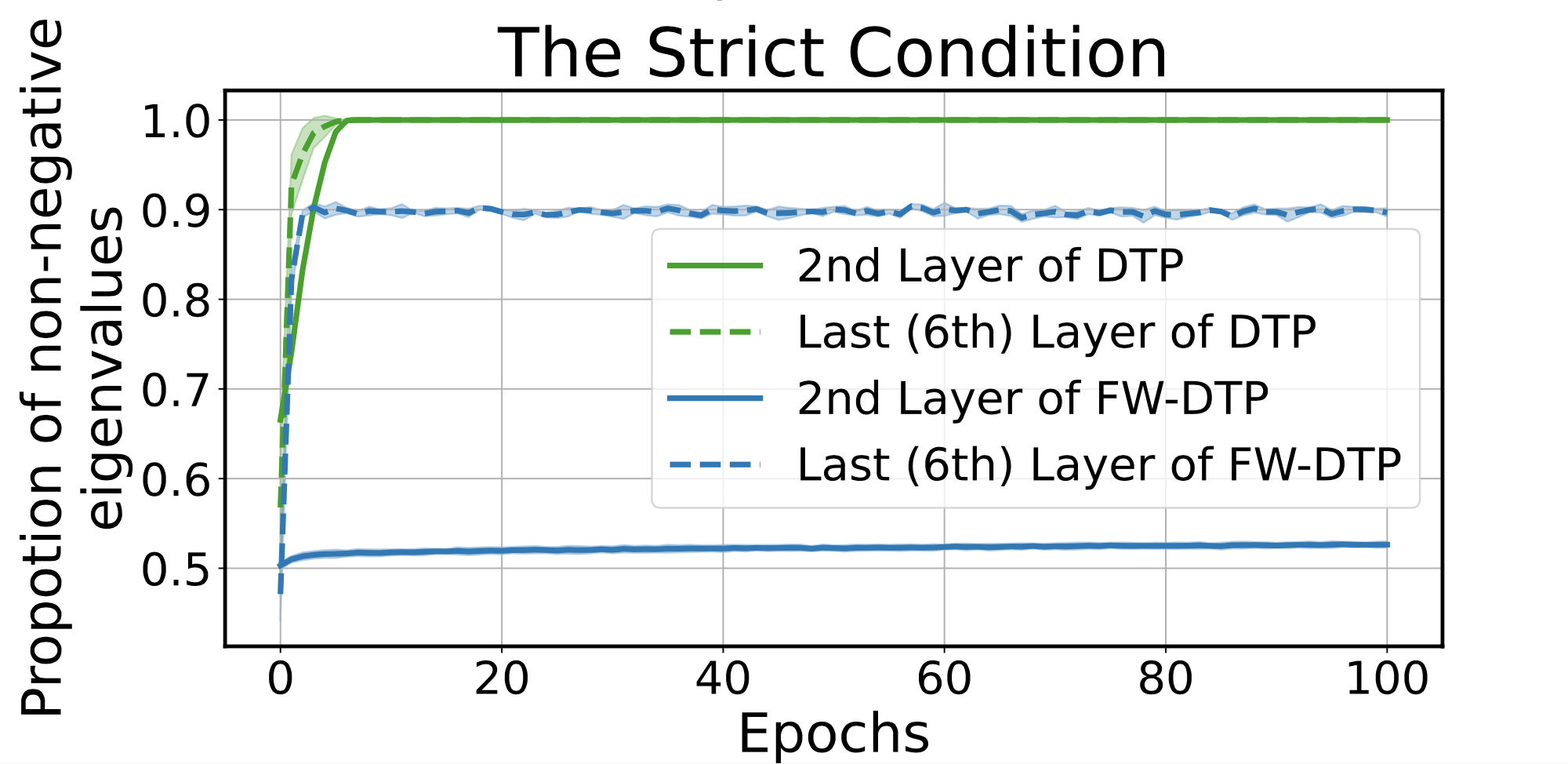}
        \subcaption{Positive semi-definiteness of $J_{f_{l}}J_{g_{l}}$}
        \label{fig:strict-condition}
      \end{minipage}
    \end{tabular}
    \caption{The Jacobian conditions of FW-DTP and DTP on MNIST with the mean and standard deviation over five different seeds. (a) Trace of $J_{f_{l}} J_{g_{l}}$ (the values of the trace on the 2nd layer of DTP are scaled by 0.1). We see that all values are positive. (b) The proportion of non-negative eigenvalues. We see the difference between DTP and FW-DTP.}
    \label{fig:condition}
 \end{figure*}

\noindent {\bf Results.} Figure~\ref{fig:condition} shows the results of the last (sixth) layer and the second layer as representatives of intermediate layers.
In Figure~\ref{fig:probabilistic-condition}, we see that the trace of $J_{f_{l}}J_{g_{l}}$ is positive from the first epoch, and is increasing during training process of DTP and FW-DTP.
In contrast, in Figure~\ref{fig:strict-condition},
we see the difference between DTP and FW-DTP.
With DTP, all eigenvalues are non-negative after the tenth epoch on both layers.
On the other hand,
with FW-DTP,
some of eigenvalues are negative.
We see that $\approx 90\%$ of eigenvalues are non-negative in the last layer,
but only $\approx 53\%$ of them are non-negative in the second layer.

These results confirm that FW-DTP satisfies only the weak condition expressed in Eq.~(\ref{eq:probabilistic-condition}) automatically
, while DTP satisfies both of the weak and strict conditions. 

\subsection{Comparison with TP and DTP Variants}\label{ex2}
\noindent {\bf Experimental set-up.}
The purpose of this experiment is to demonstrate that the performance of FW-DTP is comparable with or even better than that of DTP.
We compared image classification performance of TP \cite{tp}, DTP \cite{dtp}, DRL \cite{theoretical}, L-DRL \cite{scaling}, and FW-DTP on four datasets: MNIST~\cite{mnist}, Fashion-MNIST (F-MNIST)~\cite{fashion}, CIFAR-10 and CIFAR-100 \cite{cifar}.
Following previous studies \cite{assessing, theoretical}, a fully connected network consists of 6 layers each with 256 units was used for MNIST and F-MNIST.
Another fully connected network consists of 4 layers each with 1,024 units was used for CIFAR-10/100.
Because FW-DTP halves the number of the learnable parameters by fixing the feedback weights, we also report results with a half number of leanable parameters with DTP, DRL and L-DRL.
The activation function and the optimizer were the same as those used in \ref{ex1}. Details of the hyperparameters are provided in the appendix. 

\begin{table*}[t]
    \caption{Test error (\%) obtained on four image classification datasets reported with the mean and standard deviation over five different seeds. For the hyperparameter search, 5,000 samples from the training set are used as the validation set. The best and the second best results are marked in bold and with an underline, respectively. The columns of \#PARAMS is the number of learnable parameters (the sum of numbers of feedforward and the feedback networks).}
    \label{tab:eval}
    \begin{center}\begin{small}\begin{sc}
        \begin{tabular}{l|ccc|ccc}
            \toprule
            Methods     & \#Params & MNIST & F-MNIST & \#Params & CIFAR-10 & CIFAR-100\\
            \midrule
            BP
            & 0.5M & $1.85_{\pm 0.09}$ & $10.42_{\pm 0.08}$ & 6.3M & $46.16_{\pm 1.15}$ & $75.96_{\pm 0.52}$\\
            FA~\cite{fa}          & 0.5M & $2.94_{\pm 0.09}$ & $12.58_{\pm 0.35}$ & 6.3M & $51.33_{\pm 0.81}$ & $77.43_{\pm 0.21}$\\
            \midrule
            TP
            & 1.1M & $78.99_{\pm 2.04}$ & $-$ & 13.0M & $-$ & $-$\\
            DTP~\cite{dtp}        
            & 0.5M & $3.24_{\pm 0.15}$ & $11.86_{\pm 0.14}$ & 6.3M & $52.17_{\pm 0.79}$ & $77.89_{\pm 0.39}$\\
            & 1.1M & \underline{$2.77_{\pm 0.10}$} & \underline{$11.77_{\pm 0.16}$} & 13.0M & $52.01_{\pm 0.80}$ & $77.11_{\pm 0.20}$\\
            DRL~\cite{theoretical}
            & 0.5M & $3.13_{\pm 0.03}$ & $12.75_{\pm 0.52}$ & 6.3M & $50.11_{\pm 0.67}$ & $76.69_{\pm 0.30}$\\
            & 1.1M & $2.84_{\pm 0.09}$ & $12.15_{\pm 0.25}$ & 13.0M & \bm{$48.79_{\pm 0.58}$} &             \underline{$75.62_{\pm 0.35}$}\\
            L-DRL~\cite{scaling}
            & 0.5M & $3.14_{\pm 0.03}$ & $12.45_{\pm 0.36}$ & 6.3M & $49.58_{\pm 0.33}$ & $76.72_{\pm 0.26}$\\
            & 1.1M & $2.82_{\pm 0.10}$ & $12.29_{\pm 0.46}$ & 13.0M & $49.84_{\pm 0.55}$ & \bm{$75.62_{\pm 0.31}$}\\
            FW-DTP
            & 0.5M & \bm{$2.76_{\pm 0.10}$} & \bm{$11.76_{\pm 0.37}$} & 6.3M & \underline{$48.97_{\pm 0.32}$} & $76.76_{\pm 0.45}$\\
            \bottomrule
        \end{tabular}
    \end{sc}\end{small}\end{center}
\end{table*}

\noindent {\bf Results.}
The results are summarized in Table~\ref{tab:eval}. 
As can be seen, FW-DTP is comparable with DTP and its variants.
FW-DTP outperformed DTP in all datasets.
This supports that
FW-DTP works as a training algorithm
even if it does not satisfy the strict condition of Jacobians.
This also confirms that even with fixed feedback weights, FW-DTP propagates targets to decrease cross-entropy loss
via the feedback path with the function $g^{\nu}_{l}$ for difference correction.
Comparison with DRL and L-DRL showed some limitation of FW-DTP.
FW-DTP outperformed them on MNIST, F-MNIST, and CIFAR-10 when the number of learnable parameters was the same. 
On CIFAR-100, the test error of FW-DTP was not the best among them. 
However, 
when the number of 
parameters was the same, 
the difference in the test error between DTP and DRL or L-DRL was only $\le 0.1\%$. 
Note that the goal of this study is not to outperform them but to analyze how and why FW-DTP works as a training algorithm with empirical evidence.

\subsection{Hyperparameter Sensitivity and Computational Efficiency}\label{ex3}
Here, we investigate hyperparameter sensitivity and the computational cost of FW-DTP.

\begin{figure*}[h]
    \centering
    \hspace{-2em}
    \includegraphics[width=17cm]{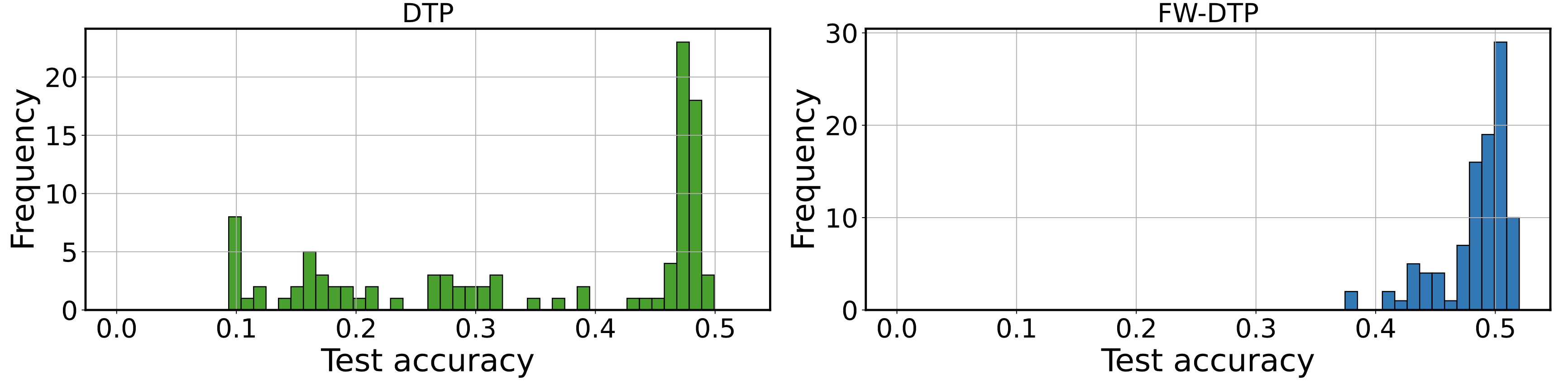}
    \caption{Histogram of test accuracies achieved under different hyperparameters on CIFAR-10.}
    \label{fig:sensitivity}
\end{figure*}

\noindent {\bf Hyperparameter sensitivity.}
We investigate how sensitive DTP and FW-DTP are to different hyperparameters. Namely, we tested 100 different random configurations.
More specifically, denoting by $\alpha \in \mathbb{R}^{H}$ the flattened hyperparameters where $H$ is the number of hyperparameters,
each $\alpha_{i}$ was randomly sampled so that
$\log (\alpha_{i}) \sim U( \log(0.2 \bar{\alpha}_{i}), \log(5 \bar{\alpha}_{i}))$ where $U$ is the uniform distribution and $\bar{\alpha}$ is the hyperparameter used in \ref{ex2}. (see Appendix \ref{app:params} for details).
The histograms of the test accuracies on CIFAR-10 are visualized in Figure~\ref{fig:sensitivity}.
As can be seen, FW-DTP is less sensitive than DTP to hyperparameters. This is because DTP needs the complicated interactions between feedforward and feedback training, as discussed in the previous work \cite{assessing}, while FW-DTP drops these complexities by relaxing the conditions of Jacobians from the strict one to the weak one.

\begin{table}[t]
    \caption{Training time [sec] per epoch of FW-DTP, DTP, DRL, L-DRL and BP on CIFAR-10.}
    \label{tab:cost}
    \begin{center}\begin{small}\begin{sc}
        \begin{tabular}{lccc}
            \toprule
            {} & Time[sec] & Ratio to FW-DTP & Error[\%]\\
            \midrule
            FW-DTP  & $2.22_{\pm 0.02}$ & $1.00_{\pm 0.00}$ & $48.97_{\pm 0.32}$\\
            DTP     & $8.32_{\pm 0.36}$ & $3.74_{\pm 0.17}$ & $52.01_{\pm 0.80}$\\
            DRL     & $9.52_{\pm 0.08}$ & $4.29_{\pm 0.05}$ & $48.79_{\pm 0.58}$\\
            L-DRL   & $8.86_{\pm 0.08}$ & $3.99_{\pm 0.05}$ & $49.84_{\pm 0.55}$\\
            BP      & $0.76_{\pm 0.03}$ & $0.34_{\pm 0.01}$ & $46.16_{\pm 1.15}$\\
            \bottomrule
        \end{tabular}
    \end{sc}\end{small}\end{center}
\end{table}

\noindent {\bf Computational Cost.} We compare the computational cost of each method on CIFAR-10 in Table~\ref{tab:cost}. 
4 GPUs (Tesla P100-SXM2-16GB) with 56 CPU cores are used to measure computational time.
For DTP, DRL and L-DRL, the feedback weights are updated five times in each iteration.
FW-DTP is $\approx$ 3.0 times slower than BP and $> 3.7$ times faster than DTP.
This shows that BP is still better in terms of computational cost, however, FW-DTP is one of the most efficient methods in DTP variants.
More detailed settings are described in the appendix.
\section{Discussion}
In this paper, we proposed FW-DTP, which fixes feedback weights during training, and experimentally confirmed that its test performance is consistently better than that of DTP on four image-classification datasets, while the hyperparameter sensitivity and the computational cost are reduced.
Further, we showed the strict and weak conditions of Jacobians, by which we explained the difference between FW-DTP and DTP. Finally, we discuss limitations and future work.

\noindent {\bf Biological plausibility.} A limitation of FW-DTP is that it does not fulfill some biological constraints such as Dale's law \cite{dale} and spiking networks \cite{spiking1,spiking2,spiking3}.
We have shown in Analysis 2 that the composite function $f_{l} \circ g_{l}$ forms a layer-wise autoencoder even
with fixed feedback weights because we have a function $g^{\nu}_{l}$ derived from difference correction.
However, allowing $g^{\nu}_{l} \not = \textit{id}$ may harm biological plausibility.
Notably, this is not a problem only for FW-DTP.
If we apply DTP to a non-injective feedforward function, a non-identity function $g^{\nu}_{l}$ often remains.
We hope our exact formulation of DTP helps researchers to analyze the behaviour of DTP in future.

\noindent {\bf Scalability.} Another limitation in this work is that all of the four datasets are for image classification and are relatively small. We chose them because of two reasons: 1) they are suitable for analyzing Jacobian matrices during training to see the difference between FW-DTP and DTP,
and
2) they are suitable for repeating many experiments with different hyper-parameters for evaluating the sensitivity.
Recently, some improved targets propagated beyond layers \cite{theoretical, scaling} perform comparable with BP on large-scale datasets.
From the point of view of fixed feedback weights, these methods may be related to the
direct feedback alignment \cite{dfa, dfa-new}. Exploring a method to add such feedback paths efficiently with some fixed feedback weights would be an interesting and necessary direction for future work.

\noindent {\bf New research direction.}
In this study, we assumed $f^{\mu}_{l} = \textit{id}$ in the decomposed encoder for a fair comparison of FW-DTP with DTP and its variants.
However, it is worth noting that exploring non-identify fixed function $f^{\mu}_{l}$,
as well as exploring different restrictions to the function space $\mathfrak{O}_{l}$ would open a new research direction.
In particular, the following symmetry in FW-DTP would be effective to explore new biologically plausible function families: $f^{\mu}_{l}, g^{\mu}_{l}$ are fixed, and $f^{\nu}_{l}, g^{\nu}_{l}$ are determined by a parameter $\theta$.
This direction includes research topics about how to fix weights in conjunction with feedback alignment methods \cite{dfa-new,fa-new,fa-condition},
and how to parameterize paired functions with some reparametrization tricks.
Under the weak condition of Jacobians, there must be fruitful function families that have never been investigated for propagating targets.
\section*{Acknowledgement}
This work was an outcome of a research project, Development of Quality Foundation for Machine-Learning Applications, supported by DENSO IT LAB Recognition and Learning Algorithm Collaborative Research Chair (Tokyo Tech.).
This work was also supported by JSPS KAKENHI Grant Number JP22H03642.
\bibliographystyle{aaai23}

\clearpage

\def\AlgoDTP{
\begingroup
\removelatexerror
    \begin{figure}[!t]
        \begin{algorithm}[H]
        \caption{Difference Target Propagation~(DTP) \cite{dtp} with the original reconstruction loss.}
        \label{fig:algo-dtp}
            \KwHyper{Learning Rate $\alpha_f$, $\alpha_b$, Stepsize $\beta$, Feedback Update Frequency $N_b$, Standard Deviation $\sigma$}\\
            \KwReq{Dataset $D$, Loss $\mathcal{L}$, Epochs $N$, Encoder $f_l$ parameterized with $W_l~(l=1\dots L)$, Decoder $g_l$ parameterized with $\Omega_l~(l=1\dots L)$}\\
            
            Pre-train only feedback network:\\
            \For{\KwAll $(x,y) \in D$}{
                $h_0 = x$\\
                \For{$l=1$ \KwTo $L$}{
                    $h_l = f_l(h_{l-1})$\\
                }
                Update the feedback weights:\\ 
                \For{$j = 1$ \KwTo $N_b$}{
                    \For{$l=1$ \KwTo $L$}{
                        $r_l = h_l + \epsilon,~\epsilon\sim N(0,\sigma I)$\\
                        $r_{l+1} = f_{l+1}(r_l)$\\
                        $r_{l}^{rec} = g_{l+1}(r_{l+1})$\\
                        $L'_l = \|r_l^{rec}-r_l \|^2_2$\\
                        $\Omega_l \leftarrow \Omega_l -\alpha_b \nabla_{\Omega_l} L'_l$\\
                    }
                }
            }
            
            Train the feedforward and feedback networks:\\
            \For{$i=1$ \KwTo $N$}{
                \For{\KwAll $(x,y) \in D$}{
                    $h_0 = x$\\
                    \For{$l=1$ \KwTo $L$}{
                        $h_l = f_l(h_{l-1})$\\
                    }
                    
                    Update the feedback weights:\\
                    \For{$j = 1$ \KwTo $N_b$}{
                        \For{$l=1$ \KwTo $L$}{
                            $r_l = h_l + \epsilon,~\epsilon\sim N(0,\sigma I)$\\
                            $r_{l+1} = f_{l+1}(r_l)$\\
                            $r_{l}^{rec} = g_{l+1}(r_{l+1})$\\
                            $L'_l = \|r_l^{rec}-r_l \|^2_2$\\
                            $\Omega_l \leftarrow \Omega_l -\alpha_b \nabla_{\Omega_l} L'_l$\\
                        }
                    }
                    
                    Compute targets:\\
                    $\tau_L = h_L - \beta \nabla_{h_L}\mathcal{L}(h_L, y)$\\
                    \For{$l=L-1$ \KwTo $1$}{
                        $\tau_l = g_{l+1}(\tau_{l+1}) + h_l - g_{l+1}(h_{l+1})$\\
                    }
                    
                    Update the forward weights:\\
                    \For{$l=1$ \KwTo $L$}{
                        $L_l = \|\tau_l-h_l\|^2_2$\\
                        $W_l \leftarrow W_l -\alpha_f \nabla_{W_l} L_l$\\
                    }
                }
            }
        \end{algorithm}
    \end{figure}
\endgroup
}

\def\AlgoDRL{
\begingroup
\removelatexerror
    \begin{figure}[!t]
        \begin{algorithm}[H]
        \caption{DTP with Difference Reconstruction Loss (DRL) \cite{theoretical}.}
        \label{fig:algo-drl}
            \KwHyper{Learning Rate $\alpha_f$, $\alpha_b$, Stepsize $\beta$, Feedback Update Frequency $N_b$, Standard Deviation $\sigma$, Tikhonov damping constant $\lambda$}\\
            \KwReq{Dataset $D$, Loss $\mathcal{L}$, Epochs $N$, Encoder $f_l$ parameterized with $W_l~(l=1\dots L)$, Decoder $g_l$ parameterized with $\Omega_l~(l=1\dots L)$}\\
            
            Pre-train only feedback network:\\
            \For{\KwAll $(x,y) \in D$}{
                $h_0 = x$\\
                \For{$l=1$ \KwTo $L$}{
                    $h_l = f_l(h_{l-1})$\\
                }
                Update the feedback weights:\\
                \For{$j = 1$ \KwTo $N_b$}{
                    \For{$l=1$ \KwTo $L$}{
                        $r_l = h_l + \epsilon,~\epsilon\sim N(0,\sigma I)$\\
                        \For{$k=l+1$ \KwTo $L$}{
                            $r_k = f_k(r_{k-1})$\\
                        }
                        $r_L^{rec} = r_L$\\
                        \For{$k=L-1$ \KwTo $l$}{
                            $r_k^{rec} = g_{k+1}(r_{k+1}^{rec}) + h_k -g_{k+1}(h_{k+1})$\\
                        }
                        $L'_l = \|r_l^{rec}-r_l \|^2_2 + \lambda \|\Omega_l\|^2_F$\\
                        $\Omega_l \leftarrow \Omega_l -\alpha_b \nabla_{\Omega_l} L'_l$\\
                    }
                }
            }
            
            Train the feedforward and feedback networks:\\
            \For{$i=1$ \KwTo $N$}{
                \For{\KwAll $(x,y) \in D$}{
                    $h_0 = x$\\
                    \For{$l=1$ \KwTo $L$}{
                        $h_l = f_l(h_{l-1})$\\
                    }
                    
                    Update the feedback weights:\\
                    \For{$j = 1$ \KwTo $N_b$}{
                        \For{$l=1$ \KwTo $L$}{
                            $r_l = h_l + \epsilon,~\epsilon\sim N(0,\sigma I)$\\
                            \For{$k=l+1$ \KwTo $L$}{
                                $r_k = f_k(r_{k-1})$\\
                            }
                            $r_L^{rec} = r_L$\\
                            \For{$k=L-1$ \KwTo $l$}{
                                $r_k^{rec} = g_{k+1}(r_{k+1}^{rec}) + h_k -g_{k+1}(h_{k+1})$\\
                            }
                            $L'_l = \|r_l^{rec}-r_l \|^2_2 + \lambda \|\Omega_l\|^2_F$\\
                            $\Omega_l \leftarrow \Omega_l -\alpha_b \nabla_{\Omega_l} L'_l$\\
                        }
                    }
                    
                    Compute targets:\\
                    $\tau_L = h_L - \beta \nabla_{h_L}\mathcal{L}(h_L, y)$\\
                    \For{$l=L-1$ \KwTo $1$}{
                        $\tau_l = g_{l+1}(\tau_{l+1}) + h_l - g_{l+1}(h_{l+1})$\\
                    }
                    
                    Update the forward weights:\\
                    \For{$l=1$ \KwTo $L$}{
                        $L_l = \|\tau_l-h_l\|^2_2$\\
                        $W_l \leftarrow W_l -\alpha_f \nabla_{W_l} L_l$\\
                    }
                }
            }
    \end{algorithm}
\end{figure}
\endgroup
}

\def\AlgoLDRL{
\begingroup
\removelatexerror
    \begin{figure}[!t]
        \begin{algorithm}[H]
        \caption{DTP with Local Difference Reconstruction Loss (L-DRL) \cite{scaling}.}
        \label{fig:algo-ldrl}
            \KwHyper{Learning Rate $\alpha_f$, $\alpha_b$, Stepsize $\beta$, Feedback Update Frequency $N_b$, Standard Deviation $\sigma$}\\
            \KwReq{Dataset $D$, Loss $\mathcal{L}$, Epochs $N$, Encoder $f_l$ parameterized with $W_l~(l=1\dots L)$, Decoder $g_l$ parameterized with $\Omega_l~(l=1\dots L)$}\\
            
            Pre-train only feedback network:\\
            \For{\KwAll $(x,y) \in D$}{
                $h_0 = x$\\
                \For{$l=1$ \KwTo $L$}{
                    $h_l = f_l(h_{l-1})$\\
                }
                Update the feedback weights:\\
                \For{$j = 1$ \KwTo $N_b$}{
                    \For{$l=1$ \KwTo $L$}{
                        $r_l = h_l + \epsilon,~\epsilon\sim N(0,\sigma I)$\\
                        $r_{l+1} = f_{l+1}(r_l)$\\
                        $r_{l}^{rec} = g_{l+1}(r_{l+1}) + h_l - g_{l+1}(h_l)$\\
                        $s_{l+1} = h_{l+1} + \eta,~\eta\sim N(0,\sigma I)$\\
                        $s_{l}^{rec} = g_{l+1}(s_{l+1}) + h_l - g_{l+1}(h_l)$\\
                        $L'_l = -(r_l-h_l)^t(r^{rec}_l-h_l) + \frac{1}{2}\|s^{rec}_l-h_l\|^2_2$\\
                        $\Omega_l \leftarrow \Omega_l -\alpha_b \nabla_{\Omega_l} L'_l$\\
                    }
                }
            }
            
            Train the feedforward and feedback networks:\\
            \For{$i=1$ \KwTo $N$}{
                \For{\KwAll $(x,y) \in D$}{
                    Propagate activations:\\
                    $h_0 = x$\\
                    \For{$l=1$ \KwTo $L$}{
                        $h_l = f_l(h_{l-1})$\\
                    }
                    
                    Update the feedback weights:\\
                    \For{$j = 1$ \KwTo $N_b$}{
                        \For{$l=1$ \KwTo $L$}{
                            $r_l = h_l + \epsilon,~\epsilon\sim N(0,\sigma I)$\\
                            $r_{l+1} = f_{l+1}(r_l)$\\
                            $r_{l}^{rec} = g_{l+1}(r_{l+1}) + h_l - g_{l+1}(h_l)$\\
                            $s_{l+1} = h_{l+1} + \eta,~\eta\sim N(0,\sigma I)$\\
                            $s_{l}^{rec} = g_{l+1}(s_{l+1}) + h_l - g_{l+1}(h_l)$\\
                            $L'_l = -(r_l-h_l)^t(r^{rec}_l-h_l) + \frac{1}{2}\|s^{rec}_l-h_l\|^2_2$\\
                            $\Omega_l \leftarrow \Omega_l -\alpha_b \nabla_{\Omega_l} L'_l$\\
                        }
                    }
                    
                    Compute targets:\\
                    $\tau_L = h_L - \beta \nabla_{h_L}\mathcal{L}(h_L, y)$\\
                    \For{$l=L-1$ \KwTo $1$}{
                        $\tau_l = g_{l+1}(\tau_{l+1}) + h_l - g_{l+1}(h_{l+1})$\\
                    }
                    
                    Update the forward weights:\\
                    \For{$l=1$ \KwTo $L$}{
                        $L_l = \|\tau_l-h_l\|^2_2$\\
                        $W_l \leftarrow W_l -\alpha_f \nabla_{W_l} L_l$\\
                    }
                }
            }
    \end{algorithm}
\end{figure}
\endgroup
}

\def\AlgoFWDTP{
\begingroup
\removelatexerror
    \begin{figure}[!t]
        \begin{algorithm}[H]
        \caption{Fixed-Weight Difference Target Propagation (FW-DTP).}
        \label{fig:algo-fwdtp}
            \KwHyper{Learning Rate $\alpha_f$, Stepsize $\beta$}\\
            \KwReq{Dataset $D$, Loss $\mathcal{L}$, Epochs $N$, Encoder $f_l$ parameterized with $W_l~(l=1\dots L)$, Decoder $g_l$ parameterized with fixed $B_l~(l=1\dots L)$}\\
            
            Train the feedforward network:\\
            \For{$i=1$ \KwTo $N$}{
                \For{\KwAll $(x,y) \in D$}{
                    $h_0 = x$\\
                    \For{$l=1$ \KwTo $L$}{
                        $h_l = f_l(h_{l-1})$\\
                    }
                    
                    Compute targets:\\
                    $\tau_L = h_L - \beta \nabla_{h_L}\mathcal{L}(h_L, y)$\\
                    \For{$l=L-1$ \KwTo $1$}{
                        $\tau_l = g_{l+1}(\tau_{l+1}) + h_l - g_{l+1}(h_{l+1})$\\
                    }
                    
                    Update the forward weights:\\
                    \For{$l=1$ \KwTo $L$}{
                        $L_l = \|\tau_l-h_l\|^2_2$\\
                        $W_l \leftarrow W_l -\alpha_f \nabla_{W_l} L_l$\\
                    }
                }
            }
    \end{algorithm}
\end{figure}
\endgroup
}

\appendix
\section{Formulation Details}
\def\appfigA{
  \begin{figure*}[t]
    \centering
    \includegraphics[width=13.0cm]{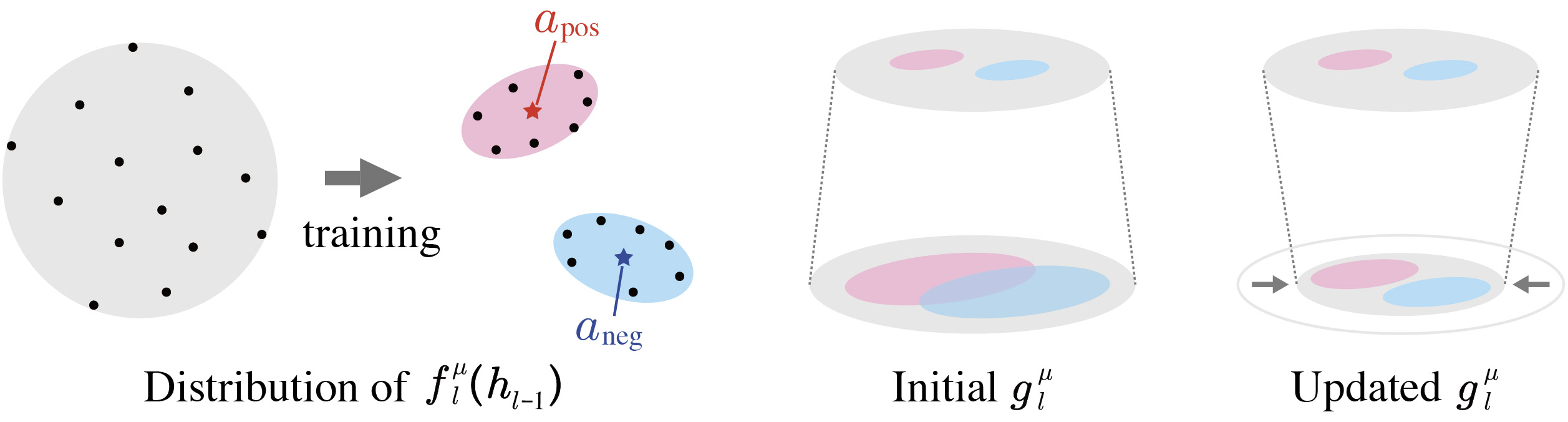}
    \caption{Distribution of $f^{\mu}_{l}(h_{l-1})$ before and after training $f^{\nu}_{l-1}$ and updates of $g^{\mu}_{l}$.}
    \label{appfigA}
  \end{figure*}
}

\def\appfigB{
  \begin{figure*}[t]
    \centering
    \includegraphics[width=14.0cm]{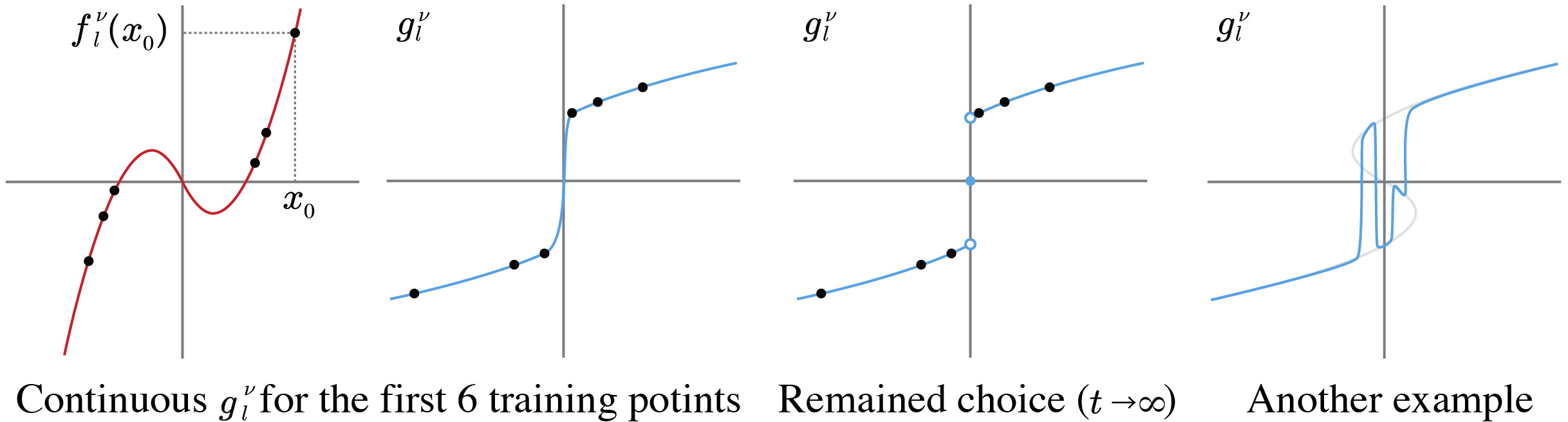}
    \caption{Example 1.}
    \label{appfigB}
  \end{figure*}
}

\def\appfigC{
  \begin{figure*}[t]
    \centering
    \includegraphics[width=14.0cm]{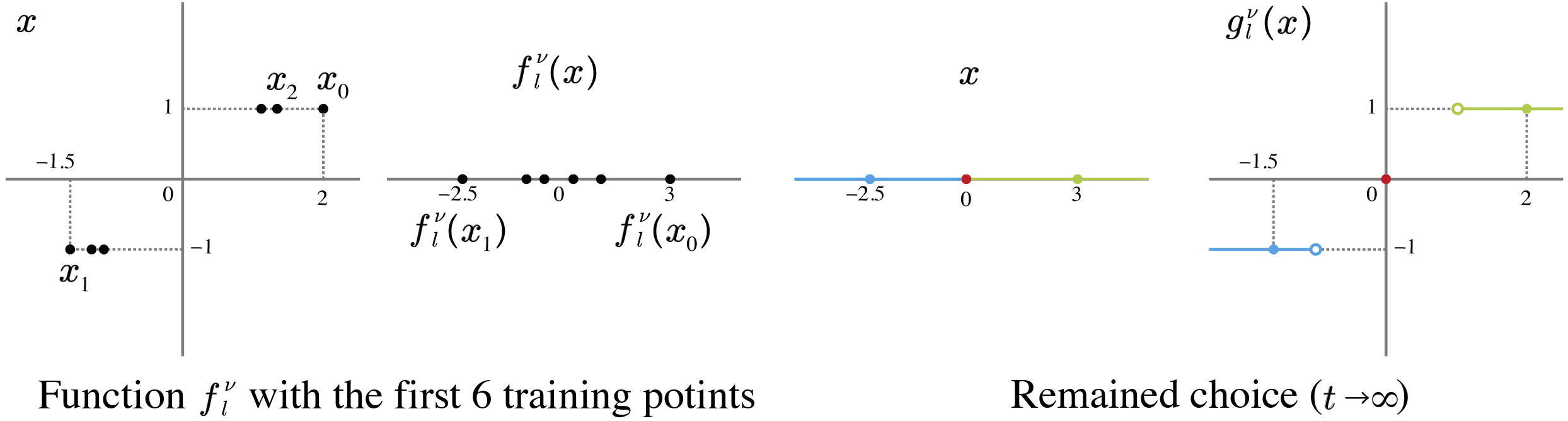}
    \caption{Example 2.}
    \label{appfigC}
  \end{figure*}
}

This section presents the details of our formulations. We show that with DTP, non-injective feedforward functions lead to non-continuous feedback functions, and discuss pros and cons of updating feedback weights. We start from the definition of training.

\def\rest#1{\left. #1 \right|}
\def\norest#1#2{#1}
\def\restA#1{\left. #1 \right|_{A^{\infty}_{i}}}

\noindent {\bf Definition A.1 (Training).}
Let $f_{l}: \mathcal{X}_{l-1} \to \mathcal{X}_{l}$ and 
$g_{l}: \mathcal{X}_{l} \to \mathcal{X}_{l-1}$ be the $l$-th encoder and decoder, respectively,
where $\mathcal{X}_{l-1}$ is a non-empty set, and $\mathcal{X}_{l} = \text{Im} f_{l}$ (i.e., the codomain of $f_{l}$ is defined so that $f_{l}$ is surjective).
Suppose they are decomposed into $f_{l} =  f^{\nu}_{l} \circ f^{\mu}_{l}$ and $g_{l} =  g^{\nu}_{l} \circ g^{\mu}_{l}$, where 
there exists two intermediate hidden spaces $\mathcal{Z}_{l}$ and $\mathcal{W}_{l}$ such that $f^{\mu}_{l} : \mathcal{X}_{l-1} \to \mathcal{Z}_{l}$,
$f^{\nu}_{l} : \mathcal{Z}_{l} \to \mathcal{X}_{l}$,
$g^{\mu}_{l} : \mathcal{X}_{l} \to \mathcal{W}_{l}$, and
$g^{\nu}_{l} : \mathcal{W}_{l} \to \mathcal{X}_{l-1}$.
Denoting by $\mathfrak{q}_{l} = (f^{\nu}_{l}, f^{\mu}_{l}, g^{\nu}_{l}, g^{\mu}_{l}) \in \mathfrak{O}_{l}$ a quadruplet of the four functions,
a {\it training step} is defined as the process to update it from $\mathfrak{q}_{l}^{(t)}$ to $\mathfrak{q}_{l}^{(t+1)}$ by
solving or approximately solving the following problem:
\mathsize\begin{align}
\mathfrak{q}^{(t+1)}_{l} = \argmin_{\mathfrak{q}_{l} \in \mathfrak{O}^{(t)}_{l}[\mathfrak{q}^{(t)}]} \mathcal{O}^{(t)}_{l}
\end{align}\normalsize
where $t \in \mathbb{N} = \{0, 1, 2, \cdots\}$ is the time step, $\mathcal{O}^{(t)}_{l}$ is the objective function, and $\mathfrak{O}^{(t)}_{l}[\mathfrak{q}^{(t)}] \subset \mathfrak{O}_{l}$ is the restricted function space (search space). Here, $\mathfrak{O}_{l}$ is the whole search space given by a product space $\mathfrak{O}_{l} = \mathcal{F}^{\nu}_{l} \times \mathcal{F}^{\mu}_{l} \times \mathcal{G}^{\nu}_{l} \times \mathcal{G}^{\mu}_{l}$ where
\mathsize\begin{align}
\mathcal{F}^{\nu}_{l} \subset \{f^{\nu}_{l} : \mathcal{Z}_{l} \to \mathcal{X}_{l}\},~
\mathcal{F}^{\mu}_{l} \subset \{f^{\mu}_{l} : \mathcal{X}_{l-1} \to \mathcal{Z}_{l}\},\nonumber\\
\mathcal{G}^{\nu}_{l} \subset \{g^{\nu}_{l} : \mathcal{W}_{l} \to \mathcal{X}_{l-1}\},~
\mathcal{G}^{\mu}_{l} \subset \{g^{\mu}_{l} : \mathcal{X}_{l} \to \mathcal{W}_{l}\}
\end{align}\normalsize
are subsets of function spaces.
\\

With this definition, the training step of target propagation (TP) is formulated as follows.

\def\pomegaspace{p_{\Omega}}
\def\pthetaspace{p_{\Theta}}
\noindent {\bf Definition A.2 (Target Propagation).}
Let $p_{\theta}$ and $q_{\omega}$ be two parameterized functions where $\theta \in \Theta$ and $\omega \in \Omega$ are parameters.
Also, define parameterized function spaces $\pomegaspace$, $\pthetaspace$ as $\pomegaspace = \{p_\omega : \omega \in \Omega\}$, $\pomegaspace = \{p_\theta : \theta \in \Theta\}$, respectively.
The training step of TP is defined with the following search space $\mathfrak{O}_{l}$, the restricted search space $\mathfrak{O}^{(t)}_{l}$, and the objective function $\mathcal{O}^{(t)}_{l}$:
\mathsize\begin{align}
\mathfrak{O}_{l} =
\{
&(f^{\nu}_{l}, f^{\mu}_{l}, g^{\nu}_{l}, g^{\mu}_{l}):\nonumber\\
&f^{\nu}_{l} \in \pthetaspace,~f^{\mu}_{l} = \textit{id},~g^{\nu}_{l} = \textit{id},~g^{\mu}_{l} \in \pomegaspace
\}
\end{align}\normalsize
where $\textit{id}$ is the identity function\footnote{This assumes $\mathcal{Z}_{l} = \mathcal{W}_{l} = \mathcal{X}_{l-1}$.}.
Given an initial quadruplet 
$\mathfrak{q}^{(0)}_{l} = (p_{\theta_{0}}, \textit{id}, \textit{id}, p_{\omega_{0}})$
with randomly chosen $\theta_{0} \in \Theta$ and $\omega_{0} \in \Omega$,
the restricted search spaces for $t = 1, 2, \cdots$ are defined as
\mathsize\begin{align}
&\mathfrak{O}^{(t)}_{l}[\mathfrak{q}^{(t)}] \nonumber\\
&=\begin{cases}
\{
(f^{\nu}_{l}, \textit{id}, \textit{id}, g^{\mu}_{l}):f^{\nu}_{l} \in \pthetaspace,~g^{\mu}_{l} = p_{\omega_{t}}
\} & (t \in \mathbb{E})\\
\{
(f^{\nu}_{l}, \textit{id}, \textit{id}, g^{\mu}_{l}):f^{\nu}_{l} = p_{\theta_{t}},~g^{\mu}_{l} \in \pomegaspace
\} & (t \in \mathbb{O})
\end{cases}
\end{align}\normalsize
where $\mathbb{E}$ is the set of even numbers, and $\mathbb{O}$ is the set of odd numbers.
This expresses that TP seeks new feedforward weights $\theta_{t+1}$ and new feedback weights $\omega_{t+1}$ alternately. The objective function is also switched to alternately apply the layer-wise local loss and reconstruction loss as
\mathsize\begin{align}
\label{app:ibjecttp}
\mathcal{O}^{(t)}_{l}
=\begin{cases}
\frac{ \| f_{l}^{\nu} \circ f^{\mu}_{l}(h_{l-1}) - \tau_{l} \|_{2}^{2}}{2\beta} & (t \in \mathbb{E})\\
\frac{\| f^{\mu}_{l}(h_{l-1}) +\epsilon - g^{\mu}_{l} \circ f^{\nu}_{l} ( f^{\mu}_{l}(h_{l-1}) + \epsilon)\|_{2}^{2}}{2} & (t \in \mathbb{O})
\end{cases}
\end{align}\normalsize
where $h_{l-1} \in \mathcal{X}_{l}$ is the feedforward activation from the layer $l-1$, $\tau_{l} \in \mathcal{X}_{l}$ is the target at the layer $l$,
$\epsilon$ is a small Gaussian noise,
and $\beta$ is a hyper-parameter.
We assumed $\mathcal{X}_{l}$ and $\mathcal{Z}_{l}$ are normed spaces.
\\

When $g^{\mu}_{l} \circ \psi_{l} \circ f^{\nu}_{l} \circ f^{\mu}_{l}$ is bijective, Difference target propagation (DTP) is formulated as follows.

\noindent {\bf Definition A.3 (Difference Target Propagation).} 
Suppose $g^{\mu}_{l} \circ \psi_{l} \circ f^{\nu}_{l} \circ f^{\mu}_{l} : \mathcal{X}_{l-1} \to \mathcal{W}_{l}$ is bijective.
With two parameterized function spaces $\pthetaspace$ and $\pomegaspace$, a training step of DTP is defined as follows. The search space is given by
\mathsize\begin{align}
\label{searchspaceofdtpa3}
\mathfrak{O}_{l} =
\{
&\mathfrak{q}_{l} = (f^{\nu}_{l}, f^{\mu}_{l}, g^{\nu}_{l}, g^{\mu}_{l}):\nonumber\\
&f^{\nu}_{l} \in \pthetaspace,~f^{\mu}_{l} = \textit{id},~g^{\mu}_{l} \in \pomegaspace,~\Pi(\mathfrak{q}_{l}) = 0
\}
\end{align}\normalsize
where $\Pi : \mathfrak{O}_{l} \to \mathbb{R}_{\geq 0}$ is {\it a restrictor} given by
\mathsize\begin{align}
\label{eq:restrictordtp}
\Pi(\mathfrak{q}_{l}) = d_{P,\mathcal{X}_{l-1}}(
&f^{\mu}_{l} \circ g_{l} \circ f^{*}_{l},\nonumber\\
&g^{\mu}_{l} \circ f^{*}_{l} + f^{\mu}_{l} - g^{\mu}_{l} \circ f_{l}).
\end{align}\normalsize
Note that $f^{*}_{l} = \psi_{l} \circ f_{l}$, and $\psi_{l}(h_{l}) = \tau_{l}$ is a shortcut as defined in Sec.~3.
Here, $d_{P, \mathcal{X}_{l}}$ a distance measure between two functions defined by
\mathsize\begin{align}
\label{distancemeasure}
d_{P, \mathcal{X}_{l-1}}(\chi, \chi') = \int_{\mathcal{X}_{l-1}} d \lambda (x) \|\chi(x) - \chi'(x)\|_{P}
\end{align}\normalsize
where $\chi$ and $\chi'$ are functions from $\mathcal{X}_{l-1}$ to $\mathcal{W}_{l}$.
Here, we assumed $(\mathcal{X}_{l-1}, \lambda)$ is a Lebesgue measurable space with a measure $\lambda$,
the two input functions in Eq.~(\ref{eq:restrictordtp}) are measurable functions, and $\mathcal{W}_{l}$ is a normed space with norm $P$.
For example, $L_{2}$ norm is reasonable if $\mathcal{W}_{l}$ is a real vector space.
At time step $t$, the restricted search space is given, in the same form as TP, as follows:
\small\begin{align}
&\mathfrak{O}^{(t)}_{l}[\mathfrak{q}^{(t)}]\nonumber\\
&=\begin{cases}
\{
(f^{\nu}_{l}, \textit{id}, g^{\nu}_{l}, g^{\mu}_{l}) \in \mathfrak{O}_{l}
:
f^{\nu}_{l} \in \pthetaspace,
g^{\mu}_{l} = p_{\omega_{t}}
\} & (t \in \mathbb{E})\\
\{
(f^{\nu}_{l}, \textit{id}, g^{\nu}_{l}, g^{\mu}_{l}) \in \mathfrak{O}_{l}
:
f^{\nu}_{l} = p_{\theta_{t}},~
g^{\mu}_{l} \in \pomegaspace
\} & (t \in \mathbb{O})
\end{cases}.
\end{align}\normalsize
The objective function is given by Eq.~(\ref{app:ibjecttp}).\\

DTP works deterministically because we have the following proposition.

\noindent {\bf Proposition A.1.}
The following search space $\bar{\mathfrak{O}}$ is {\it deterministic}, i.e., $\forall \mathfrak{q}, \mathfrak{q'} \in \bar{\mathfrak{O}}~\mathfrak{q} = \mathfrak{q'}~\text{a.e.}$
\footnote{Given two quadruplets $\mathfrak{q} = (f^{\nu}, f^{\mu}, g^{\nu}, g^{\mu})$ and $\mathfrak{q}' = (f'^{\nu}, f'^{\mu}, g'^{\nu}, g'^{\mu})$, this means that $f^{\nu} = f'^{\nu}$, $f^{\mu} = f'^{\mu}$, $g^{\nu} = g'^{\nu}~\text{a.e.}$, and $g'^{\mu} = g^{\mu}$. If we ignore function differences on zero-measure sets, this is equivalent to $|\bar{\mathfrak{O}}| = 1$. We call the space is deterministic because of this.}, if functions $p_{1}, p_{2}, p_{3}$ and $\psi_{l}$ are bijective:
\mathsize\begin{align}
\label{searchspacebar}
\bar{\mathfrak{O}} =
\{
&\mathfrak{q}_{l} = (f^{\nu}_{l}, f^{\mu}_{l}, g^{\nu}_{l}, g^{\mu}_{l}) \in \mathfrak{O}_{l}:\nonumber\\
&f^{\nu}_{l} = p_{1},~f^{\mu}_{l} = p_{2},~g^{\mu}_{l} = p_{3},~\Pi(\mathfrak{q}_{l}) = 0
\}
\end{align}\normalsize
where $\Pi$  is given by Eq.~(\ref{eq:restrictordtp}).

(Proof) 
From $\Pi(\mathfrak{q}_{l}) = 0$, we have
\mathsize\begin{align}
\label{eq:propa1eq1}
f^{\mu}_{l} \circ g^{\nu}_{l} \circ g^{\mu}_{l} \circ f^{*}_{l}
=
g^{\mu}_{l} \circ f^{*}_{l}
+
f^{\mu}_{l}
-
\xgb{l} \circ f_{l}~~\lambda\text{-a.e.}.
\end{align}\normalsize
Since there exists inverse functions
$(f^{\nu}_{l})^{-1}, (\psi_{l})^{-1}, (g^{\mu}_{l})^{-1}$ and $(f^{\mu}_{l})^{-1}$,
we have the following function that satisfies Eq.~(\ref{eq:propa1eq1}):
\mathsize\begin{align}
g^{\nu}_{l} &= (f^{\mu}_{l})^{-1} \circ (g^{\mu}_{l} \circ f^{*}_{l}+f^{\mu}_{l} -\xgb{l} \circ f_{l})\nonumber\\
&~~~~~~  \circ (f^{*}_{l})^{-1} \circ (g^{\mu}_{l})^{-1} \nonumber\\
&= (f^{\mu}_{l})^{-1} \circ ( \textit{id} + f^{\mu}_{l} \circ (f^{*}_{l})^{-1} \circ (g^{\mu}_{l})^{-1} \nonumber\\
&~~~~~~ -\xgb{l} \circ f_{l} \circ (f^{*}_{l})^{-1} \circ (g^{\mu}_{l})^{-1}) \nonumber\\
&= (f^{\mu}_{l})^{-1} \circ ( \textit{id} + (f^{\nu}_{l})^{-1} \circ (\psi_{l})^{-1} \circ (g^{\mu}_{l})^{-1} \nonumber\\
&~~~~~~ -\xgb{l} \circ (\psi_{l})^{-1} \circ (g^{\mu}_{l})^{-1}).
\end{align}\normalsize
This is the unique function that satisfies $\Pi(\mathfrak{q}_{l}) = 0$ with exception of values on zero-measure sets because the inverse of a bijective function is unique.
Thus, we have 
\mathsize\begin{align}
\bar{\mathfrak{O}} =
\{
&\mathfrak{q}_{l} = (f^{\nu}_{l}, f^{\mu}_{l}, g^{\nu}_{l}, g^{\mu}_{l}):\nonumber\\
&f^{\nu}_{l} = p_{1},~f^{\mu}_{l} = p_{2},~g^{\nu}_{l} \in \mathfrak{G},~g^{\mu}_{l} = p_{3}\}
\end{align}\normalsize
where $\mathfrak{G} = \{g: g = \xfbcheck{l} ~\lambda''\text{-a.e.}\}$, $\lambda''$ is a measure on $\mathcal{W}_{l}$, and
\mathsize\begin{align}
\label{eqgcheck}
\xfbcheck{l}
:=
(f^{\mu}_{l})^{-1} \circ 
(
\textit{id}
&+ (f^{\nu}_{l})^{-1} \circ (\psi_{l})^{-1} \circ (g^{\mu}_{l})^{-1}\nonumber \\
&- \xgb{l} \circ (\psi_{l})^{-1} \circ (g^{\mu}_{l})^{-1}
).
\end{align}\normalsize
This shows that $\bar{\mathfrak{O}}$ is deterministic.\\

Practically, DTP and its variants often outperform TP because the restrictor $\Pi$ is well designed to satisfy some requirements of learning as discussed in Sec.~3.5.
However, as pointed out in recent studies, DTP is unstable if feedforward function is not injective.
With our formulation,
the unstability is realized by that 
$\bar{\mathfrak{O}}$ becomes empty.

\noindent {\bf Proposition A.2.}
Suppose $p_{1} \circ p_{3} : \mathcal{X}_{l-1} \to \mathcal{X}_{l}$ is not injective, and $p_{2}$ and $\psi_{l}$ are bijective for the search space $\bar{\mathfrak{O}}$ in Eq.~(\ref{searchspacebar}).
Suppose also $(\mathcal{X}_{l-1}, \lambda)$ and $(\mathcal{X}_{l}, \lambda')$ are Lebesgue measurable spaces. If there exists two subsets $O_{1}, O_{2} \subset \mathcal{X}_{l-1}$ that satisfy
\mathsize\begin{align}
&\lambda (O_{1} \cap O_{2}) = 0,\\
&\lambda' ( p_{1} \circ p_{3} (O_{1}) \cap p_{1} \circ p_{3} (O_{2})) > 0,
\end{align}\normalsize
the search space is empty, i.e., $|\bar{\mathfrak{O}}| = 0$.

(Proof)
Assume, for contradiction, that $|\bar{\mathfrak{O}}| > 0$.
For $\mathfrak{q}_{l} = (f^{\nu}_{l}, f^{\mu}_{l}, g^{\nu}_{l}, g^{\mu}_{l}) \in \bar{\mathfrak{O}}$, from $\Pi(\mathfrak{q}_{l}) = 0$, we have
\begin{align}
\label{eq:proof1}
&f^{\mu}_{l} \circ g^{\nu}_{l} \circ g^{\mu}_{l} \circ f^{*}_{l}(o_{1}) \nonumber\\
&= g^{\mu}_{l} \circ f^{*}_{l}(o_{1}) + f^{\mu}_{l}(o_{1}) - \xgb{l} \circ f_{l}(o_{1})~\lambda\text{-a.e.~} o_{1} \in O_{1}
\end{align}\normalsize
and
\begin{align}
\label{eq:proof2}
&f^{\mu}_{l} \circ g^{\nu}_{l} \circ g^{\mu}_{l} \circ f^{*}_{l}(o_{2})\nonumber\\
&= g^{\mu}_{l} \circ f^{*}_{l}(o_{2}) + f^{\mu}_{l}(o_{2})- \xgb{l} \circ f_{l}(o_{2})~\lambda\text{-a.e.~} o_{2} \in O_{2}.
\end{align}\normalsize
From $\lambda' ( p_{1} \circ p_{3} (O_{1}) \cap p_{1} \circ p_{3} (O_{2})) > 0$, we have a positive-measure pair set
\mathsize\begin{align}
Q = \{(o_{1}, o_{2}): p_{1} \circ p_{3} (o_{1}) = p_{1} \circ p_{3} (o_{2})\}
\end{align}\normalsize
where $Q \subset O_{1} \times O_{2}$ and each of $(o_{1}, o_{2}) \in Q$ satisfies
\mathsize\begin{align}\label{eq:proof3}
f^{\mu}_{l} \circ g^{\nu}_{l} \circ g^{\mu}_{l} \circ f^{*}_{l}(o_{1})
=&
f^{\mu}_{l} \circ g^{\nu}_{l} \circ g^{\mu}_{l} \circ \psi_{l} \circ p_{1} \circ p_{3} (o_{1})\nonumber\\
=&
f^{\mu}_{l} \circ g^{\nu}_{l} \circ g^{\mu}_{l} \circ \psi_{l} \circ p_{1} \circ p_{3} (o_{2})\nonumber\\
=&
f^{\mu}_{l} \circ g^{\nu}_{l} \circ g^{\mu}_{l} \circ f^{*}_{l}(o_{2}).
\end{align}\normalsize
Further, the pairs also satisfies
\mathsize\begin{align}
\label{eq:proof4}
&g^{\mu}_{l} \circ f^{*}_{l}(o_{1}) + f^{\mu}_{l}(o_{1}) - \xgb{l} \circ f_{l}(o_{1}) \nonumber \\
&= g^{\mu}_{l} \circ f^{*}_{l}(o_{2}) + f^{\mu}_{l}(o_{1}) - \xgb{l} \circ f_{l}(o_{2}).
\end{align}\normalsize
From Eqs.~(\ref{eq:proof1}), (\ref{eq:proof2}), (\ref{eq:proof3}), and (\ref{eq:proof4}), we have
\mathsize\begin{align}
&g^{\mu}_{l} \circ f^{*}_{l}(o_{2})+f^{\mu}_{l}(o_{1})-\xgb{l} \circ f_{l}(o_{2})\nonumber\\
&=g^{\mu}_{l} \circ f^{*}_{l}(o_{2})+f^{\mu}_{l}(o_{2})-\xgb{l} \circ f_{l}(o_{2})
\end{align}\normalsize
and thus $f^{\mu}_{l}(o_{1}) = f^{\mu}_{l}(o_{2})$ follows for paris in $Q$.
Since $f^{\mu}_{l}$ is bijective, the measure of
$\{o_{1} : (o_{1}, o_{2}) \in Q, o_{1} = o_{2}\}$ is positive. This contradicts to the assumption of $\lambda(O_{1} \cap O_{2}) = 0$.\\

This raises a question: How is DTP working with non injective feedforward function?
The answer is that DTP applies the restrictor $\Pi$ at each time step by
relaxing the Eq.~(\ref{distancemeasure}) as
\mathsize\begin{align}
&d_{P, A_{t}}(\rest{\chi}_{A_{t}}, \rest{\chi'}_{A_{t}})\nonumber\\
&:= \int_{A_{t}} d\kappa(x) \|\rest{\chi}_{A_{t}}(x) - \rest{\chi'}_{A_{t}}(x)\|_{P}.
\end{align}\normalsize
Here, $A_{t} \subset \mathcal{X}_{l-1}$ is a measurable subset such that $\left.(g^{\mu}_{l} \circ \psi_{l} \circ f^{\nu}_{l} \circ f^{\mu}_{l})\right|_{A_{t}} : A_{t} \to \text{Im}A_{t}$ is bijective, where the notation $\rest{\chi}_{A}$ denotes restriction\footnote{We also restrict the codomain of $\chi$ to $\text{Im} A$.} of $\chi$ to $A$.
Note that $\kappa$ is a measure on $A_{t}$\footnote{This may not be the same measure as $\lambda$ on $\mathcal{X}_{l}$, even if $A_{t}$ is a zero-measure set with the measure $\lambda$, $\kappa(A_{t})$ may be positive.}.
The following gives the formulation.

\noindent {\bf Definition A.4 (DTP with non injective functions).}
A training step of DTP is defined as follows.
With two parameterized function spaces $\pthetaspace$ and $\pomegaspace$, the search space is given by
\mathsize\begin{align}
\mathfrak{O}_{l} =
\{
&(f^{\nu}_{l}, f^{\mu}_{l}, g^{\nu}_{l}, g^{\mu}_{l}):\nonumber\\
&f^{\nu}_{l} \in \pthetaspace,~f^{\mu}_{l} = \textit{id}, \nonumber\\
&g^{\nu}_{l} \in \{g : \mathcal{W}_{l} \to \mathcal{X}_{l-1}\},~g^{\mu}_{l} \in \pomegaspace
\}.
\end{align}\normalsize
At time step $t$, a {\it local restrictor} $\Pi_{t}$ is applied to determine $g^{\mu}_{l}$ as follows:
\mathsize\begin{align}
&\mathfrak{O}^{(t)}_{l}[\mathfrak{q}^{(t)}] \nonumber \\
&=\begin{cases}
\{
\mathfrak{q}_{l} = 
(f^{\nu}_{l}, \textit{id}, g^{\nu}_{l}, g^{\mu}_{l}) \in \mathfrak{O}_{l}:\\
~~f^{\nu}_{l} \in \pthetaspace,~g^{\mu}_{l} = p_{\omega_{t}},~\Pi_{t}(\mathfrak{q}_{l}) = 0
\} & (t \in \mathbb{E})\\
\{
\mathfrak{q}_{l} = 
(f^{\nu}_{l}, \textit{id}, g^{\nu}_{l}, g^{\mu}_{l}) \in \mathfrak{O}_{l}:\\
~~f^{\nu}_{l} = p_{\theta_{t}},~g^{\mu}_{l} \in \pomegaspace \Pi_{t}(\mathfrak{q}_{l}) = 0
\} & (t \in \mathbb{O})
\end{cases}
\end{align}\normalsize
where $\Pi_{t}$ is a restrictor given by
\mathsize\begin{align}
\Pi_{t}(\mathfrak{q}) = d_{P, A_{t}}(
&\rest{(f^{\mu}_{l} \circ g_{l} \circ f^{*}_{l})}_{A_{t}},\nonumber\\
&\rest{(g^{\mu}_{l} \circ f^{*}_{l})}_{A_{t}} + \rest{f^{\mu}_{l}}_{A_{t}}-\rest{(g^{\mu}_{l} \circ f_{l})}_{A_{t}})
\end{align}\normalsize
with a subset $A_{t} \subset \mathcal{X}_{l-1}$
such that $\left.(g^{\mu}_{l} \circ \psi_{l} \circ f^{\nu}_{l} \circ f^{\mu}_{l})\right|_{A_{t}}$ is bijective.
\\

If $A_{t}$ exists, we see with this definition that 
\mathsize\begin{align}
\bar{\mathfrak{O}}^{(t)}_{l}
:=\{
&\mathfrak{q}_{l} = (f^{\nu}_{l}, \textit{id}, g^{\nu}_{l}, g^{\mu}_{l}) \in \mathfrak{O}_{l}:\nonumber\\
&f^{\nu}_{l} = p_{\theta_{t}},~g^{\mu}_{l} = p_{\omega_{t}},~\Pi_{t}(\mathfrak{q}_{l}) = 0
\}
\end{align}\normalsize
becomes not empty.
More specifically, similar to Prop.~A.1,
we obtain
\mathsize\begin{align}\label{gmufordtpnoninjective}
\check{g}^{\nu (t)}_{l} = 
& (\rest{f^{\mu}_{l}}_{A_{t}})^{-1} \circ ( \textit{id}\nonumber\\
&+ (\rest{f^{\nu}_{l}}_{B_{t}})^{-1} \circ (\rest{\psi_{l}}_{C_{t}})^{-1} \circ (\rest{g^{\mu}_{l}}_{D_{t}})^{-1}\nonumber\\
&- \rest{\xgb{l}}_{D_{t}} \circ (\rest{\psi_{l}}_{C_{t}})^{-1} \circ (\rest{g^{\mu}_{l}}_{D_{t}})^{-1})
\end{align}\normalsize
where $B_{t} = \text{Im} \rest{f^{\mu}_{l}}_{A_{t}} \subset \mathcal{Z}_{l}$,
$C_{t} = \text{Im} \rest{f^{\nu}_{l}}_{B_{t}} \subset \mathcal{X}_{l}$, and
$D_{t} = \text{Im} \rest{\psi_{l}}_{C_{t}} \subset \mathcal{X}_{l}$.
Note that the domain of $\check{g}^{\nu (t)}_{l}$ is $E_{t} = \text{Im} \rest{g^{\mu}_{l}}_{D_{t}} \subset \mathcal{W}_{l}$.
Thus, we have
\mathsize\begin{align}
\bar{\mathfrak{O}}^{(t)}_{l}=\{
&\mathfrak{q}_{l} = (f^{\nu}_{l}, \textit{id}, g^{\nu}_{l}, g^{\mu}_{l}):\nonumber\\
&f^{\nu}_{l} = p_{\theta_{t}},~g^{\nu}_{l} \in \mathfrak{G}_{t},~g^{\mu}_{l} = p_{\omega_{t}}
\}
\end{align}\normalsize
where
\mathsize\begin{align}
\mathfrak{G}_{t}
= \{
\check{g} \in \{\mathcal{W}_{l} \to \mathcal{X}_{l}\}
:
\rest{g}_{E_{t}} =
\check{g}^{\nu (t)}_{l}
\},
\end{align}\normalsize
that is, a function in $\mathfrak{G}_{t}$ is an extension of $\check{g}^{\nu (t)}_{l}$ from $E_{t}$ to $\mathcal{W}_{t}$.
At each time step $t$ in DTP, there must be one function quadruplet implicitly chosen from $\bar{\mathfrak{O}}^{(t)}_{l}$.
Intuitive understanding of this is that DTP determines $g^{\nu}_{l}$ by only seeing a subset $A_{t} \subset \mathcal{X}_{l-1}$ where every function becomes bijective.

The subset $A_{t}$ is determined depends on the input sequence for training.
In online learning, where inputs are given point-wise, $A_{t}$ always exists. The following proposition is trivial, but helps to understand what DTP does.

\noindent {\bf Proposition A.3.}
Given an input\footnote{The input to layer $l$ is the activation $h_{l-1}$, but we use notation $x_{t}$ for simplicity.} of online learning $x_{t} \in \mathcal{X}_{l-1}$ at time $t$ to layer $l$, there exists $A_{t} \supset \{x_{t}\}$ that makes $\left.(g^{\mu}_{l} \circ \psi_{l} \circ f^{\nu}_{l} \circ f^{\mu}_{l})\right|_{A_{t}}$ bijective.

(Proof) Let $X_{t} = \{x_{t}\}$. Clearly, $A_{t} = X_{t}$ makes all functions
$\rest{f^{\mu}_{l}}_{A_{t}}$, 
$\rest{f^{\nu}_{l}}_{B_{t}}$, 
$\rest{g^{\mu}_{l}}_{C_{t}}$, and
$\rest{g^{\nu}_{l}}_{D_{t}}$
bijective because we have $|B_{t}| = |C_{t}| = |D_{t}| = 1$.\\

Piratically, in mini-batch learning, the input is given by $X_{t} = \{x_{t,i}\}_{i=1}^{N} \subset \mathcal{X}_{l-1}$.
Under some proper assumptions on $\mathcal{X}_{l-1}, \mathcal{X}_{l}$ (such as both of them are dense vector spaces whose dimension is large enough), $\left.(g^{\mu}_{l} \circ \psi_{l} \circ f^{\nu}_{l} \circ f^{\mu}_{l})\right|_{A_{t}}$ becomes bijective with $A_{t} = X_{t}$ with probability of 1.0 because $X_{t}$ is a finite zero-measure set.
This shows how DTP is working with non injective feedforward function.

From above, we obtain a sequence of the restricted spaces $\bar{\mathfrak{O}}^{(0)}_{l}, \bar{\mathfrak{O}}^{(1)}_{l}, \bar{\mathfrak{O}}^{(2)}_{l}, \cdots$,
from each of which DTP implicitly chooses one function quadruplet $\mathfrak{q} \in \bar{\mathfrak{O}}^{(t)}_{l}$ at each time step $t$.
Our interest lies in
whether there exits a common solution over $t$,
i.e., whether $\exists t^{\dag} \geq 0 \text{~s.t.~} |\bigcap_{t=t^{\dag}}^{\infty} \bar{\mathfrak{O}}^{(t)}_{l}| > 0$.
The time step $t^{\dag}$ is a large enough number to see the final phase of training where $f^{\mu}_{l}, f^{\nu}_{l}$ and $g^{\mu}_{l}$ have converged\footnote{We assumed the convergence of these three functions to analyze $g^{\nu}_{l}$. In general, they may oscillate.}.

Here, we consider online learning where $X_{t} = \{x_{t}\}$ is given as an input at time $t$, and fix $f^{\mu}_{l}, f^{\nu}_{l}$ and $g^{\mu}_{l}$ to analyze $g^{\nu}_{l}$.
The following proposition is also trivial, but shows there often exists global $g^{\nu}_{l}$.

\noindent {\bf Proposition A.4.} 
Let $X_{t} = \{x_{t}\}~(t=0,1,2,\cdots)$ be a input sequence for online training, and 
\mathsize\begin{align}
\bar{\mathfrak{O}}^{(t)}_{l} =
\{
&\mathfrak{q}_{l} = 
(f^{\nu}_{l}, f^{\mu}_{l}, g^{\nu}_{l}, g^{\mu}_{l}) \in \mathfrak{O}_{l}:\nonumber\\
&f^{\nu}_{l} = p_{1},~f^{\mu}_{l} = p_{2},~g^{\mu}_{l} = p_{3},~\Pi_{t}(\mathfrak{q}_{l}) = 0
\}
\end{align}\normalsize
be the restricted search space at time step $t$ where $A_{t} = X_{t}$ is used to define $\Pi_{t}$.
With $A = \{x_{t}\}_{t=1}^{\infty}$,
if $\left.(g^{\mu}_{l} \circ \psi_{l} \circ f^{\nu}_{l} \circ f^{\mu}_{l})\right|_{A}$ is bijective, $\bigcap_{t=t^{\dag}}^{\infty} \bar{\mathfrak{O}}^{(t)}_{l}$ is not empty.

(Proof)
Because $A$ is a finite or countably infinite set, there exists a common discrete measure for all $A_{t}$ and $A$ that satisfies
\mathsize\begin{align}
\sum_{t=1}^{\infty} \Pi_{t}(\mathfrak{q}_{l})
&= \sum_{t=1}^{\infty} d_{P, A_{t}}(
\rest{(f^{\mu}_{l} \circ g_{l} \circ f^{*}_{l})}_{A_{t}},\nonumber\\
&~~~~~~ \rest{(g^{\mu}_{l} \circ f^{*}_{l})}_{A_{t}} + \rest{f^{\mu}_{l}}_{A_{t}}-\rest{(g^{\mu}_{l} \circ f_{l})}_{A_{t}})\nonumber\\
&= d_{P, A}(
\rest{(f^{\mu}_{l} \circ g_{l} \circ f^{*}_{l})}_{A},\nonumber\\
&~~~~~~ \rest{(g^{\mu}_{l} \circ f^{*}_{l})}_{A} + \rest{f^{\mu}_{l}}_{A}-\rest{(g^{\mu}_{l} \circ f_{l})}_{A}).
\end{align}\normalsize
Replacing $A_{t}$ by $A$ in Eq.~(\ref{gmufordtpnoninjective}), we have a quadruplet $\mathfrak{q}_{l}$ that satisfies $\Pi_{t}(\mathfrak{q}_{l}) = 0$ for all $t$, and thus $\bigcap_{t=0}^{\infty} \bar{\mathfrak{O}}^{(t)}_{l}$ is not empty.

Again, since inputs for training are given sequentially, and a training set is finite or countably infinite, $\left.(g^{\mu}_{l} \circ \psi_{l} \circ f^{\nu}_{l} \circ f^{\mu}_{l})\right|_{A}$ often becomes bijective under proper assumptions.
This is the reason why DTP is applicable even if feedforward function is not injective.

However, a non-injective feedforward function may lead to non-continuous feedback function in DTP.
More specifically, the obtained $\check{g}^{\nu}_{l}$ of $\mathfrak{q}_{l} = (f^{\nu}_{l}, f^{\mu}_{l}, \check{g}^{\nu}_{l}, g^{\mu}_{l}) \in \bigcap_{t=0}^{\infty} \bar{\mathfrak{O}}^{(t)}_{l}$
may be non-continuous even if $f^{\nu}_{l}$, $f^{\mu}_{l}$, $g^{\mu}_{l}$, and $\psi_{l}$ are continuous.
Note that if $f^{\nu}_{l}$, $f^{\mu}_{l}$, $g^{\mu}_{l}$, and $\psi_{l}$ are bijective and continuous, from Eq.~(\ref{eqgcheck}), there exists continuous $\check{g}^{\nu}_{l}$, because the inverse of a continuous bijective function is continuous.
Therefore, the feedback non-continuity is caused by the feedforward non-injectiveness. The following small extreme examples help to imagine the shape of feedback functions in DTP.

\appfigB
\appfigC
\noindent {\bf Example 1.} The first example is on the one dimensional real space $\mathbb{R}$. Suppose $f^{\mu}_{l}(x) = x$, $f^{\nu}_{l}(x) = \frac{1}{2}(x^{3} - 3x)$,
$g^{\mu}_{l}(x) = x$, and inputs for training are given by $x_{t} = (-1)^{t}(1 + (t+1)^{-1})$ for $t = 0, 1, 2, \cdots$.
There is no continuous function quadruplet remains in $\bigcap_{t=0}^{\infty} \bar{\mathfrak{O}}^{(t)}_{l}$.
Figure~\ref{appfigB} illustrates one of remaining choices for
$g^{\nu}_{l}$ of $\mathfrak{q} \in \bigcap_{t=0}^{\infty} \bar{\mathfrak{O}}^{(t)}_{l}$ given by
\mathsize\begin{align}
g^{\nu}_{l}(x) =
\begin{cases}
(x + (x^{2} - 1)^{\frac{1}{2}})^{\frac{1}{3}} \\
~~~~ + (x - (x^{2} - 1)^{\frac{1}{2}})^{\frac{1}{3}} & (|x| \geq 1)\\
\cos ( \arccos (\frac{x}{3}) ) &(0 < x < 1)\\
\cos ( \arccos (\frac{x + 2 \pi}{3})) & (-1 < x < 0)\\
0 & (x = 0)\\
\end{cases}.
\end{align}\normalsize
This is not continuous at $0$.
In the figure, we also show another example of $g^{\nu}_{l}$ where training inputs are distributed around $x=0$. This is an extreme case but DTP may propagate targets in similar ways.

\noindent {\bf Example 2.} The second example is from $\mathbb{R}^{2}$ to $\mathbb{R}$.
Suppose $f^{\mu}_{l} = \textit{id}$, $f^{\nu}_{l}(x^{(1)}, x^{(2)}) = x^{(1)} + x^{(2)}$ for $(x^{(1)}, x^{(2)}) \in \mathbb{R}^{2}$,
$g^{\mu}_{l} = \textit{id}$, and inputs are given by
$(x^{(1)}_{t}, x^{(2)}_{t}) = ((-1)^{t}(1 + (t+1)^{-1},(-1)^{t})$ for $t = 0, 1, 2, \cdots$.
There is no continuous function quadruplet remains in $\bigcap_{t=0}^{\infty} \bar{\mathfrak{O}}^{(t)}_{l}$.
Figure~\ref{appfigC} illustrates $g^{\nu}_{l} : \mathbb{R} \to \mathbb{R}^{2}$ given by
\mathsize\begin{align}
g^{\nu}_{l}(x) =
\begin{cases}
(x-1,1) & (x > 0)\\
(x+1,-1) & (x < 0)\\
(0,0) & (x = 0)
\end{cases}.
\end{align}\normalsize
This shows that feedforward function with dimension reduction may also cause non-continuous feedback in DTP.
\\

Finally, we discuss pros and cons of updating parameters of $g^{\mu}_{l}$ with a reconstruction loss.
The advantages are clear.
If the feedforward function is bijective, 
$g^{\mu}_{l}$ approaches to the inverse of $f^{\nu}_{l}$ and this makes better targets.
Note that $\check{g}^{\nu}_{l}$ approaches the inverse of $f^{\mu}_{l}$, which is assumed to be the identity function in DTP.
Recent studies have shown that modifying reconstruction loss significantly improve the performance of DTP \cite{drtp, theoretical} by propagating information from some top layers ($l' > l$) to $l$. These techniques contribute to stabilizing training.

A disadvantage we found here is that
updating $g^{\mu}_{l}$ potentially
makes it difficult to find continuous $g^{\nu}_{l}$.
As we can image from the two extreme examples, the feedback function easily becomes non-Lipschitz continuous function, and may approach to non-continuous function.
\appfigA
Figure~\ref{appfigA} considers a binary classification problem for discussion.
At the beginning of training, $f^{\mu}_{l-1}$ is a randomly chosen function, and thus
the minimum radius that covers the inputs $A = \{x_{t}\}_{t=0}^{\infty} \subset \mathcal{X}_{l-1}$
for layer $l$  is large (these inputs are the activations from $(l-1)$-th layer).
After some training steps, $f_{l-1}$ often approaches to non-injective function ${f}^{\star}_{l-1} : \mathcal{X}_{l-2} \to \{a_{\text{neg}}, a_{\text{pos}}\}$, which maps all points in $\mathcal{X}_{l-2}$ into two points $a_{\text{neg}}, a_{\text{pos}} \in \mathcal{X}_{l-1}$, where $a_{\text{neg}}, a_{\text{pos}}$ represent the positive class and the negative class, respectively\footnote{This assumption is strong, but a $C$-class classification problem can be seen as a problem to find an non-injective mapping from inputs to $\{1, 2, \cdots, C\}$. Thus, forward functions often approaches to some non-injective functions to some extent.}.
In this case, we only have points around $a_{\text{pos}}$ and $a_{\text{neg}}$ to compute reconstruction loss to update $g^{\mu}_{l}$.
According to Prop.~A.2., this potentially
 narrows the search space because
 the overlap of two images $g^{\mu}_{l}(O_{1}) \cap g^{\mu}_{l}(O_{2})$ of two disjoint sets $O_{1}, O_{2} \subset \mathcal{X}_{l}$
 tends to increase around $a_{\text{pos}}$ and/or $a_{\text{neg}}$ by updating $g^{\mu}_{l}$.
 Unfortunately, adding small $\epsilon$ to each $x_{t}$ to compute reconstruction loss does not fully resolves this problem. Updating $f^{\mu}_{l}$ also has similar effect, but this would be unavoidable.
 
 From this discussion we reach to an idea to slowdown updates of feedback weights.
For example, in Eq.~(\ref{app:ibjecttp}), we can replace $\mathbb{O}$ and $\mathbb{E}$ by
$\mathbb{N}_{n} := \{nk : k \in \mathbb{N}\} = \{0, n, 2n, 3n, \cdots \}$ and $\mathbb{N}^{c}_{n} = \mathbb{N} \setminus \mathbb{N}_{n}$, respectively, with $n > 2$, as follows:
\mathsize\begin{align}
\mathcal{O}^{(t)}_{l} =
\begin{cases}
\frac{ \| f_{l}^{\nu}(h_{l-1}) - \tau_{l} \|_{2}^{2}}{2\beta} & (t \in \mathbb{N}^{c}_{n})\\
\frac{ \| h_{l-1} +\epsilon - g^{\mu}_{l} \circ f^{\nu}_{l} (h_{l-1} + \epsilon)\|_{2}^{2}}{2} & (t \in \mathbb{N}_{n})
\end{cases}.
\end{align}\normalsize
FW-DTP is the algorithm obtained from $n \to \infty$, where there exists reconstruction loss, but it will never be used.

Biological plausibility need to be discussed carefully, because we decomposed encoder and decoder into two functions respectively.
The main idea of TP is
to parameterize $f^{\nu}_{l}$ and $g^{\mu}_{l}$ in the same way, e.g., $f^{\nu}_{l}(x) = \sigma(W x)$ and $g^{\mu}_{l}(x) = \sigma(\Omega x)$ where $W, \Omega$ are learnable matrices and $\sigma$ is an activation function.
However, with our formulation, the decoder
$g_{l} = g^{\nu}_{l} \circ g^{\mu}_{l}$ has $g^{\nu}_{l} \in \mathfrak{G}_{l}$, 
in contrast to $f^{\mu}_{l} = \textit{id}$ for the encoder.
FW-DTP did not resolve this asymmetry, but it has another symmetry that $(f^{\mu}_{l}, g^{\mu}_{l})$ are fixed and $(f^{\nu}_{l}, g^{\nu}_{l})$ are determined by a parameter $\theta$.

This would help to explore new variants of DTP.
In this paper, we fixed $f^{\mu}_{l} = \textit{id}$ to fairly compare FW-DTP with TP and DTP variants. However, in future work, designing new search space, the the following generalized DTP is interesting:

\noindent {\bf Definition A.5 Generalized DTP (GDTP).}
Let $\pthetaspace$ be a parameterized function space.
GDTP uses the search space of
\mathsize\begin{align}
\mathfrak{O}_{l} =
\{
&\mathfrak{q}_{l} = (h_{l}, f^{\nu}_{l}, f^{\mu}_{l}, g^{\nu}_{l}, g^{\mu}_{l}):\nonumber\\
&h_{l} \in \pthetaspace,~f^{\mu}_{l} = \{r_{1}\},~f^{\nu}_{l} \in \{\mathcal{Z}_{l} \to \mathcal{X}_{l+1}\},\nonumber \\
&g^{\mu}_{l} = \{r_{2}\},~g^{\nu}_{l} \in \{\mathcal{W}_{l} \to \mathcal{X}_{l}\},~\Pi^{*}(\mathfrak{q}_{l}) = 0
\}
\end{align}\normalsize
where $h_{l}$ is parameterized, $f^{\mu}_{l}, g^{\mu}_{l}$ are fixed functions $r_{1}$ and $r_{2}$,
and $\Pi^{*}$ is a restrictor that determines $f^{\mu}_{l}$ and $g^{\nu}_{l}$.

When $f^{\nu}_{l} = h_{l}$ and $f^{\mu}_{l} = \textit{id}$, GDTP reduces DTP, but for biological plausibility, 
(1) $f^{\nu}_{l}$ and $g^{\nu}_{l}$ should have the same form,
(2) $r_{1}$ and $r_{2}$ should have the same form,
and
(3) $\Pi^{*}$ should be symmetric to $(f^{\nu}_{l}$, $g^{\nu}_{l})$ and $(f^{\mu}_{l}$, $g^{\mu}_{l})$. We left this for future work.

\section{Proof of Eq.~(\ref{eq:probabilistic-condition}) $\Leftrightarrow$ Eq.~(\ref{eq:trace})}\label{app:proof}
Let $\mathcal{N}(0, \mathbb{I})$ be the standard Gaussian distribution. We have
\begin{eqnarray*}
    \mathbb{E}_{\epsilon \sim \mathcal{N} (0, \mathbb{I})} \left[\epsilon^{\top} J_{f_{l}}J_{g_{l}} \epsilon\right]
    &=& \mathbb{E}_{\epsilon \sim \mathcal{N} (0, \mathbb{I})} \left[\mathrm{tr}(\epsilon\epsilon^{\top} J_{f_{l}}J_{g_{l}}) \right]\\
    &=& \mathrm{tr}\left(\mathbb{E}_{\epsilon \sim \mathcal{N} (0, \mathbb{I})} \left[\epsilon\epsilon^{\top} J_{f_{l}}J_{g_{l}} \right]\right)\\
    &=& \mathrm{tr}\left(\mathbb{E}_{\epsilon \sim \mathcal{N} (0, \mathbb{I})} \left[\epsilon\epsilon^{\top} \right]J_{f_{l}}J_{g_{l}}\right)\\
    &=& \mathrm{tr}\left(\mathbb{I}\hspace{2pt} J_{f_{l}}J_{g_{l}}\right)\\
    &=& \mathrm{tr}\left(J_{f_{l}}J_{g_{l}}\right)
\end{eqnarray*}
and thus we have
\mathsize\begin{align}
    \mathbb{E}_{\epsilon \sim \mathcal{N} (0, \mathbb{I})}\left[\epsilon^{\top} J_{f_{l}}J_{g_{l}} \epsilon\right]  \ge 0
    \Leftrightarrow
    \mathrm{tr}(J_{f_{l}}J_{g_{l}}) \ge 0.
\end{align}\normalsize

\begin{table*}[t!]
    \caption{Test error ($\%$) obtained with and without batch normalization on four image classification datasets reported with the mean and standard deviation over five different seeds. The best results are marked in bold.}
    \label{tab:bn-all}
    \begin{center}\begin{small}\begin{sc}
        \begin{tabular}{l|cccc}
        \toprule
        Methods & MNIST & F-MNIST & CIFAR-10 & CIFAR-100\\
        \midrule
        DTP w/o BN      & $2.77_{\pm 0.10}$ & $11.77_{\pm 0.16}$ & $52.01_{\pm 0.80}$ & $77.11_{\pm 0.20}$\\
        DTP w BN        & $3.51_{\pm 0.17}$ & $12.74_{\pm 0.22}$ & $52.58_{\pm 0.43}$ & $77.36_{\pm 0.31}$\\
        FW-DTP w/o BN   & $2.86_{\pm 0.14}$ & $13.49_{\pm 0.60}$ & $51.22_{\pm 2.03}$ & \bm{$75.38_{\pm 0.34}$}\\
        FW-DTP w BN     & \bm{$2.76_{\pm 0.10}$} & \bm{$11.76_{\pm 0.37}$} & \bm{$48.97_{\pm 0.32}$} & $76.76_{\pm 0.45}$\\
        \bottomrule
        \end{tabular}
    \end{sc}\end{small}\end{center}
\end{table*}
\section{Necessity and Effectiveness of Batch Normalization in DTP and FW-DTP}\label{app:bn}
In this section, we investigate the necessity and effectiveness of batch normalization (BN) \cite{bn} in DTP-derived methods.
First, we overview BN.
Let $\mathcal{B}=\{h_i\}_{i=1}^{M}$ be a mini-batch of activation, where $M$ is the mini-batch size and $h_{i} = (h^{(1)}_i,\cdots, h^{(D)}_i)$ is a $D$-dimensional activation vector. 
The normalized activation $\hat{h}_i$ is defined by
\mathsize\begin{align}
    \hat{h}_i^{(d)} = \frac{h_i^{(d)}-\mu_\mathcal{B}^{(d)}}{\sigma_\mathcal{B}^{(d)}}
\end{align}\normalsize
where
\mathsize\begin{align}
    \mu_\mathcal{B}^{(d)} = \frac{1}{M}\sum_{j=1}^M h_j^{(d)},~
    \sigma_\mathcal{B}^{(d)} = \frac{1}{M}\sum_{j=1}^M (h_j^{(d)}-\mu_\mathcal{B}^{(d)})^2
\end{align}\normalsize
for $d=1,\cdots D$.
The batch normalization function $\text{BN}_{\gamma, \delta}(\cdot)$ is defined by
\mathsize\begin{align}
    \text{BN}_{\gamma, \delta}(h_i) := \gamma \hat{h}_i + \delta
\end{align}\normalsize
where $\gamma$ and $\delta$ are learnable parameters. 

In FW-DTP, the weak condition expressed in Eq.~(\ref{eq:probabilistic-condition}) between the feedforward and feedback networks is acquired in the leaning process; however, in the early stage of the learning process, the target $\tau_l$ propagated by the feedback network deviates significantly from the activation $h_l$ if BN is not applied, which destabilizes the learning process and makes the test performance worse. BN resolves this optimization problem. We fix the parameters $\gamma,~\delta$ and use $\gamma=1,~\delta=0$ for all hidden layers in both of the feedforward and feedback networks. Using this fixed BN, the mean of each dimension of the targets computed according to Eq.~(\ref{eq:dtp-propagation}) are also $0$, so $h_l$ and $\tau_l$ are relatively close to each other from the beginning of the learning process. Note that a constant other than $0$ can be used as the fixed value of $\gamma$, while we chose $0$ for the simplicity and symmetry with respect to origin.

The test performances obtained by DTP and FW-DTP with and without BN are shown in Table~\ref{tab:bn-all}. 
In FW-DTP, the test performances are considerably improved by using BN for all datasets except CIFAR-100. On the other hand, in DTP, the test performances are slightly worse with BN than without for all datasets.
This suggests that fixing the mean and variance has a negative effect; since the feedback network is trained in parallel in DTP, the target $\tau_l$ and the feedforward activation $h_l$ are close even without BN.

From these results, it can be concluded that FW-DTP without BN can achieve almost the same level of test performance as DTP, and it performs even better when BN is used; however, BN is not so effective for DTP.

\section{Connection between FW-DTP and DTP}
In this section, we discuss the connection between FW-DTP and DTP in terms of learning rate for feedback network $\alpha_b$. If $\alpha_b$ is set to zero in DTP, the feedback weights are not updated, so the algorithm becomes the same as FW-DTP. What then is the behavior of the test performance when $\alpha_b$ is asymptotically approaching zero?

\begin{figure}[t]
    \hspace{-1em}
    \centering
    \includegraphics[width=8cm]{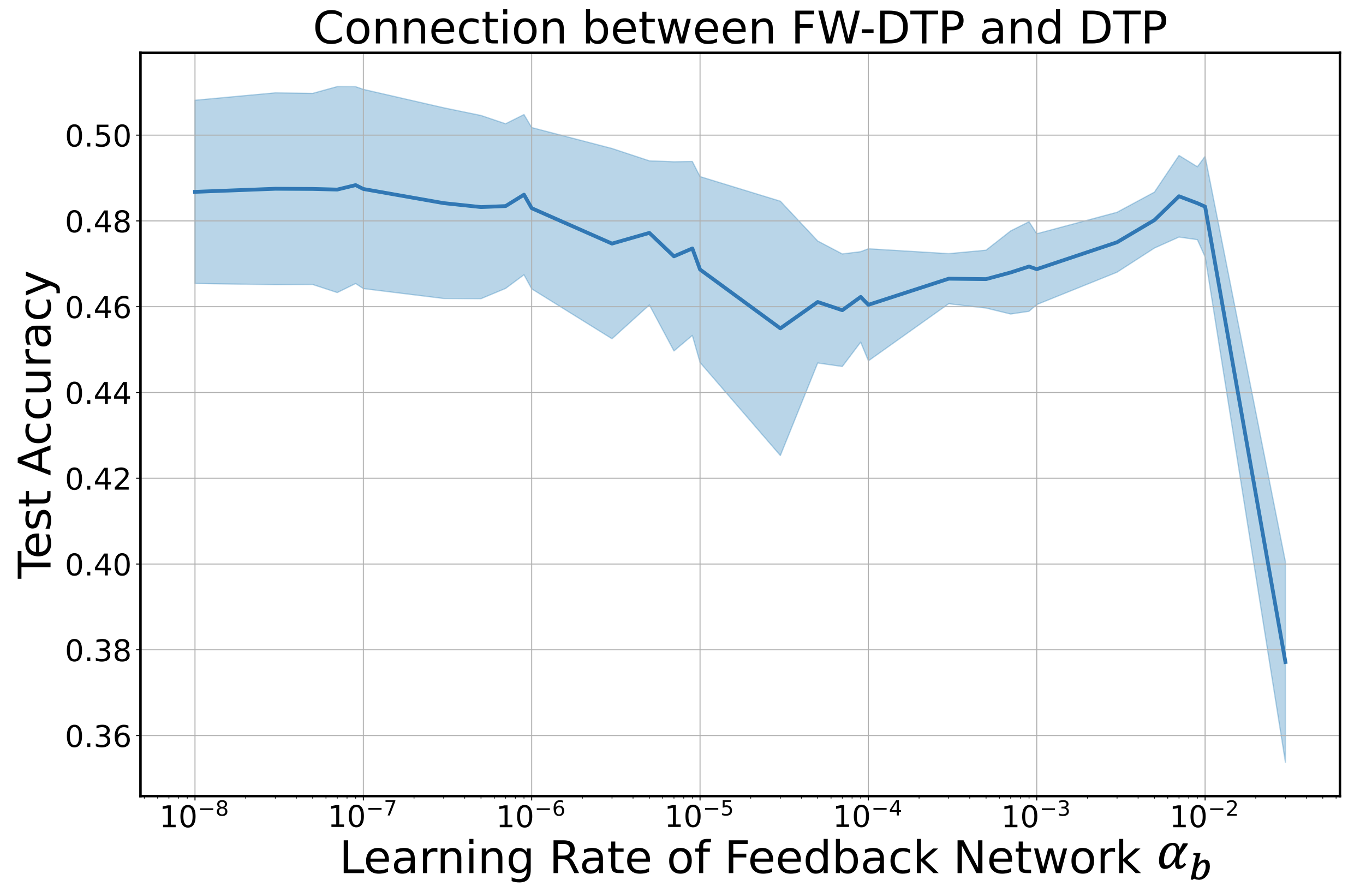}
    \caption{The connection between FW-DTP and DTP by adjusting the learning rate for the feedback network on CIFAR-10. Test accuracy are shown with the mean and standard deviation over ten different seeds. The values of hyperparameters other than the learning rate of feedback network are the same as those used in \ref{ex2}.}
    \label{fig:connection}
\end{figure}

Figure~\ref{fig:connection} shows the result on CIFAR-10. The values of hyperparameters other than $\alpha_b$ are the same as those of DTP used in \ref{ex2} and BN is not used in the activation functions. If $\alpha_b$ becomes close to 0, it asymptotically approaches FW-DTP (without BN). At $\alpha_b=0$, the test accuracy is actually $48.80\pm 2.06$, indicating that DTP converges to FW-DTP.
As the figure shows, the test accuracy reaches a local maximum around $\alpha_b=6\times 10^{-3}$, and the accuracy decreases as $\alpha_b$ is reduced.
Interestingly, the test accuracy begins to increase from around $\alpha_b=3\times 10^{-5}$, and finally, the test accuracy is on the same level as the local maximum around $\alpha_b=6\times 10^{-3}$ with a sufficiently small $\alpha_b$. Note that FW-DTP is slightly destabilized and the standard deviation is increased because BN is not used.

This result indicates that $\alpha_b$, which cannot make the feedback network a precise inverse of the feedforward network but is large enough to add some perturbation to the feedback weights, adds noise to the feedback weights and degrades the test performance. On the other hand, if $\alpha_b$ is sufficiently small, the update of feedback weights can be ignored, {\em i.e.}, DTP can approximate FW-DTP, thus improving the test performance.

\begin{table*}[t]
    \caption{Error rate against the number of layers on CIFAR 10. The number of hidden units is $1024$.}
    \label{tab:evaldepth}
    \begin{center}\begin{small}\begin{sc}
    \begin{tabular}{lccccc}
        \toprule
        \# of Layers & 5 & 6 & 7 & 8 & 9\\
        \midrule
        BP     & $48.02_{\pm 1.39}$ & $46.71_{\pm 0.70}$ & $45.24_{\pm 0.54}$ & $45.34_{\pm 0.28}$ & $46.78_{\pm 3.63}$\\
        \midrule
        DTP    & $51.66_{\pm 0.85}$ & $52.48_{\pm 0.52}$ & $51.08_{\pm 0.69}$ & $51.59_{\pm 0.57}$ & $52.03_{\pm 0.64}$\\
        FW-DTP & $\noindent {\bf 49.47_{\pm 0.28}}$ & $\noindent {\bf 50.06_{\pm 0.53}}$ & $\noindent {\bf 50.54_{\pm 0.49}}$ & $\noindent {\bf 50.50_{\pm 0.44}}$ & $\noindent {\bf 50.51_{\pm 0.10}}$\\
        \bottomrule
    \end{tabular}
    \end{sc}\end{small}\end{center}
\end{table*}

\begin{table*}[t]
    \caption{Error rate against the number of hidden units on CIFAR 10. The number of layers is $4$.}
    \label{tab:evalwidth}
    \begin{center}\begin{small}\begin{sc}
    \begin{tabular}{lccccccc}
        \toprule
        \# of Hidden Units & 64 & 128 & 256 & 512 & 1024 & 1536 & 2048\\
        \midrule
        BP     & $52.50_{\pm 0.41}$ & $51.02_{\pm 0.51}$ & $50.05_{\pm 0.41}$ & $49.58_{\pm 0.93}$ & $46.43_{\pm 1.59}$ & $44.45_{\pm 0.35}$ & $43.88_{\pm 0.69}$\\
        \midrule
        DTP    & $55.63_{\pm 0.21}$ & $54.34_{\pm 0.34}$ & $53.05_{\pm 0.35}$ & $52.14_{\pm 0.71}$ & $51.91_{\pm 0.42}$ & $52.77_{\pm 0.86}$ & $52.39_{\pm 0.50}$\\
        FW-DTP & $\noindent {\bf 53.96_{\pm 0.51}}$ & $\noindent {\bf 51.86_{\pm 0.20}}$ & $\noindent {\bf 50.53_{\pm 0.74}}$ & $\noindent {\bf 49.42_{\pm 0.42}}$ & $\noindent {\bf 49.11_{\pm 0.11}}$ & $\noindent {\bf 48.92_{\pm 0.25}}$ & $\noindent {\bf 48.77_{\pm 0.35}}$\\
        \bottomrule
    \end{tabular}
    \end{sc}\end{small}\end{center}
\end{table*}

\section{Evaluation with Large and Small Network}
In this section, we discuss the scalability of FW-DTP. Tables~\ref{tab:evaldepth} and \ref{tab:evalwidth} report accuracy of BP, DTP, and FW-DTP on CIFAR 10 with varying numbers of layers and hidden units, respectively. We see that FW-DTP consistently outperforms DTP. However, the performance gap from BP increases as the size of network increases. This shows a limitation of FW-DTP.
To deal with the scalability problem, some feedback connections from the top layer as proposed in direct feedback alignment \cite{dfa, dfa-new} and improved target propagation \cite{theoretical, scaling} would be needed. 

\begin{table*}[t!]
\centering
\caption{Comparison of characteristics.}
\begin{tabular}{l|cccccc}
\toprule
& BP & FA & DFA & TP & DTP & FW-DTP\\
\midrule
No weight transport & & $\checkmark$ & $\checkmark$ & $\checkmark$ & $\checkmark$ & $\checkmark$\\
Local error signals & & & $\checkmark$ & $\checkmark$ & $\checkmark$ & $\checkmark$\\
Function symmetry & & & & $\checkmark$ & $\checkmark$ & $\checkmark$\\
Fixed-feedback weights & & $\checkmark$ & $\checkmark$ & & & $\checkmark$\\
\bottomrule
\end{tabular}
\label{tab:algorithm_comparison}
\end{table*}

\begin{figure*}[t!]
    \centering
    \includegraphics[width=15cm]{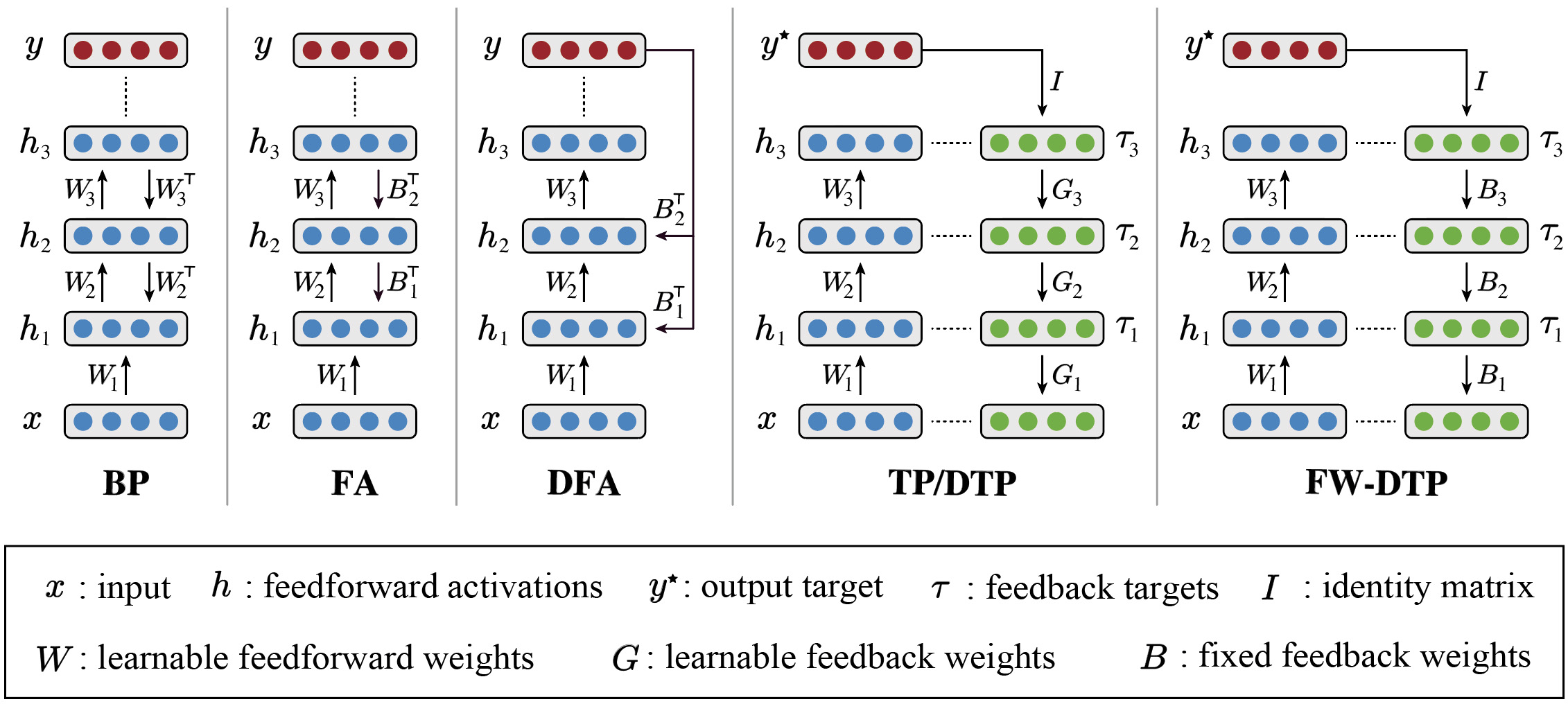}
    \caption{Comparison of propagation configurations.}
    \label{fig:algorithm_comparison}
\end{figure*}

\section{Comparison of Algorithms}
Table~\ref{tab:algorithm_comparison} summarizes characteristics of backpropagation (BP), feedback alignment (FA), direct feedback alignment (DFA) \cite{dfa}, target propagation (TP), difference target propagation (DTP), and fixed-weight difference target propagation (FW-DTP).
In terms of biological plausibility, all algorithms except of BP resolves the weight transport problem.
The error signals are local with DFA and all TP algorithms and this helps local (layer-wise) learning.
TP algorithms can have the same function on the feedforward and feedback paths.
Fixed feedback weights are used with FA, DFA, and FW-DTP.

Figure~\ref{fig:algorithm_comparison} compares propagation configurations of these algorithms. We see that FW-DTP propagates signals in a different way from FA while they commonly fix the feedback weights.

\section{Detailed Settings of Experiments}\label{app:params}
This section reports the detailed settings of the experiments in Section 4.

\noindent {\bf Algorithms.}
We first explain the exact algorithm of FW-DTP, DTP, DRL and L-DRL in Algorithms \ref{fig:algo-fwdtp}-\ref{fig:algo-ldrl}. 
For DTP, DRL and L-DRL, training is start with 1 epoch of feedback weights training, then feedforward weights and feedback weights are trained altogether with $N_b$ feedback weight updates per batch; however, FW-DTP has no pre-training of feedback weights since it completely fixes the feedback weights. 
For all methods, the encoder $f_l$ is parameterized as $f_l(h_{l-1}) = \sigma_l(W_l h_{l-1})$, where the non-linear activation function $\sigma_l$ is a hyperbolic tangent function for DTP, DRL and L-DRL. For FW-DTP, the activation function $\sigma_l(h_{l-1})$ is $\text{BN}_{1,0}(\text{tanh}(h_{l-1}))$ as explained in Appendix \ref{app:bn}. We chose the hyperbolic tangent function because it gave the best experimental results as consistently reported in the previous research \cite{dtp, assessing, theoretical}.
The decoder $g_l$ is also parameterized as $g_l(\tau_l)=\sigma_l(\Omega_l \tau_l)$, where the activation function $\sigma_l$ is the same function as the encoder's activation function and for FW-DTP, $\Omega_l=B_l$ is a fixed random matrix. Note that Algorithms \ref{fig:algo-fwdtp}-\ref{fig:algo-ldrl} show the case of the batch size is $1$ for clarity.

\AlgoFWDTP
\AlgoDTP
\AlgoDRL
\AlgoLDRL

\noindent {\bf Architectures.}
We used a stack of fully connected layers and a softmax layer as the network architecture for all methods and datasets.  More detailed architectures used for each dataset are shown below:
\begin{itemize}
    \item For MNIST: 5 fully connected layers with 256 hidden units + 1 output softmax layer with 10 units. 
    \item For Fashion-MNIST: 5 fully connected layers with 256 hidden units + 1 output softmax layer with 10 units.
    \item For CIFAR-10: 3 fully connected layers with 1024 hidden units + 1 output softmax layer with 10 units.
    \item For CIFAR-100: 3 fully connected layers with 1024 hidden units + 1 output softmax layer with 100 units.
\end{itemize}
These architectures are reported as the architectures suited to DTP in the previous research \cite{assessing}. The reduced architectures whose number of learnable parameters are halved from the architectures above were used in \ref{ex2}, and their details are shown bellow:
\begin{itemize}
    \item For MNIST: 5 fully connected layers with 164 hidden units + 1 output softmax layer with 10 units. 
    \item For Fashion-MNIST: 5 fully connected layers with 164 hidden units + 1 output softmax layer with 10 units.
    \item For CIFAR-10: 3 fully connected layers with 632 hidden units + 1 output softmax layer with 10 units.
    \item For CIFAR-100: 3 fully connected layers with 631 hidden units + 1 output softmax layer with 100 units.
\end{itemize}

\noindent {\bf Hyperparameters.}
We report the search sets of hyperparameters used in \ref{ex2}.
Table~\ref{tab:hyper-dtp-all} shows the search sets of the learning rate for the feedforward network $\alpha_f$, the stepsize $\beta$ and the learning rate for the feedback network $\alpha_b$ for DTP, DRL and L-DRL. These search sets were determined by a rough tuning of 10-fold intervals and used for all datasets.
\begin{table*}[!t]
    \caption{Search sets of hyperparameters of DTP, DRL and L-DRL for all datasets.}
    \label{tab:hyper-dtp-all}
    \begin{center}\begin{small}\begin{sc}
        \begin{tabular}{c|c}
            \toprule
            Hyperparameter & Search set\\
            \midrule
            Learning Rate $\alpha_{f}$ & $\{0.1, 0.2, 0.4, 0.8, 1, 2, 4, 8\}$\\
            Stepsize $\beta$ & $\{0.001, 0.002, 0.004, 0.008, 0.01, 0.02, 0.04, 0.08\}$\\
            Learning Rate $\alpha_{b}$ & $\{0.0001, 0.0002, 0.0004, 0.0008, 0.001, 0.002, 0.004, 0.008\}$\\
            \bottomrule
        \end{tabular}
    \end{sc}\end{small}\end{center}
\end{table*}

\begin{table*}[!t]
    \caption{Search sets of hyperparameters of FW-DTP for MNIST and F-MNIST.}
    \label{tab:hyper-fwdtp-mnist}
    \begin{center}\begin{small}\begin{sc}
        \begin{tabular}{c|c}
            \toprule
            Hyperparameter & Search set\\
            \midrule
            Learning Rate $\alpha_{f}$ & $\{0.1, 0.2, 0.4, 0.8, 1, 2, 4, 8\}$\\
            Stepsize $\beta$ & $\{0.001, 0.002, 0.004, 0.008, 0.01, 0.02, 0.04, 0.08\}$\\
            \bottomrule
        \end{tabular}
        \end{sc}\end{small}\end{center}
\end{table*}

\begin{table*}[t!]
    \caption{Search sets of hyperparameters of FW-DTP for CIFAR-10/100.}
    \label{tab:hyper-fwdtp-cifar}
    \begin{center}\begin{small}\begin{sc}
        \begin{tabular}{c|c}
            \toprule
            Hyperparameter & Search set\\
            \midrule
            Learning Rate $\alpha_{f}$ & $\{0.01, 0.02, 0.04, 0.08, 0.1, 0.2, 0.4, 0.8\}$\\
            Stepsize $\beta$ & $\{0.001, 0.002, 0.004, 0.008, 0.01, 0.02, 0.04, 0.08\}$\\
            \bottomrule
        \end{tabular}
    \end{sc}\end{small}\end{center}
\end{table*}

\begin{table*}[!t]
    \caption{The best hyperparameters used in \ref{ex1} and \ref{ex2}.}
    \label{tab:used-all}
    \begin{center}\begin{small}\begin{sc}
        \begin{tabular}{ll|cccc}
            \toprule
            Dataset & Hyperparameter & DTP
            & DRL 
            & L-DRL 
            & FW-DTP\\
            \midrule
            MNIST & Learning Rate $\alpha_{f}$
            & $4$& $4$& $4$& $0.1$\\
            {} & Stepsize $\beta$
            & $0.04$& $0.04$& $0.04$& $0.04$\\
            {} & Learning Rate $\alpha_{b}$  
            & $0.002$& $0.0002$& $0.0002$& $-$\\
            \midrule
            F-MNIST & Learning Rate $\alpha_{f}$
            & $1$& $4$& $2$& $1$\\
            {} & Stepsize $\beta$
            & $0.04$& $0.008$& $0.02$& $0.004$\\
            {} & Learning Rate $\alpha_{b}$  
            & $0.002$& $0.002$& $0.001$& $-$\\
            \midrule
            CIFAR-10 & Learning Rate $\alpha_{f}$
            & $4$& $2$& $0.2$& $0.02$\\
            {} & Stepsize $\beta$
            & $0.002$& $0.008$& $0.04$& $0.01$\\
            {} & Learning Rate $\alpha_{b}$  
            & $0.004$& $0.004$& $0.0004$& $-$\\
            \midrule
            CIFAR-100 & Learning Rate $\alpha_{f}$
            & $1$& $0.8$& $2$& $0.2$\\
            {} & Stepsize $\beta$
            & $0.01$& $0.02$& $0.008$& $0.01$\\
            {} & Learning Rate $\alpha_{b}$  
            & $0.008$& $0.002$& $0.0001$& $-$\\
            \bottomrule
        \end{tabular}
    \end{sc}\end{small}\end{center}
\end{table*}
\noindent The feedback update frequency $N_b$, standard deviation $\sigma$ and batch size $M$ were fixed to reduce tuning costs, and we used $N_b=5$, $\sigma=0.01$ and $M=256$. Note that, while the learning rates are sensitive to choice, the standard deviation had little effect on learning even when the value was set to 10x or 0.1x.
For DRL, the Tikhonov damping constant $\lambda$ was also fixed to $0$, {\em i.e.,} no regularization was applied.
For FW-DTP, the search sets of the learning rate for the feedforward network $\alpha_f$ and the stepsize $\beta$ used for MNIST and F-MNIST are shown in Table.~\ref{tab:hyper-fwdtp-mnist}. These search sets are same as those of DTP, but for CIFAR10/100, a different search sets were obtained as a result of the abovementioned rough tuning and shown in Figure~\ref{tab:hyper-fwdtp-cifar}.

For DTP, DRL and L-DRL, we initialized the weights of the feedforward networks with orthogonal matrices for all datasets. For FW-DTP, we also used orthogonal matrices to initialize the weights of the feedforward network; however, the weights of the feedback network were initialized with a uniform distribution $U(-10^{-2},10^{-2})$.

The best hyperparameters used in \ref{ex2} are shown in Table~\ref{tab:used-all}. These best hyperparameters were also used in \ref{ex1}. To find these best hyperparameters, grid search was used; 5000 samples from training set were used as a validation set, and the set of hyperparameters which obtained the best test performance at $100$ epoch were chosen as the best hyperparameters.

\noindent {\bf Hyperparameter Sampling in Sec. \ref{ex3}.}
Tables~\ref{tab:sampling-dtp} and \ref{tab:sampling-fwdtp} show sampled hyperparameters used in the evaluation of hyperparameter sensitivity in Section~4.3.
The number of hyperparameters $H$ is three for DTP and two for FW-DTP.
Note that for a fair comparison,
each hyperparameter $\gamma$ was randomly sampled so that
$\log (\gamma) \sim U(\log(0.2 \bar{\gamma}), \log(5 \bar{\gamma}))$ where $U$ is the uniform distribution and $\bar{\gamma}$ is the best hyperparameter used in Section~\ref{ex2}. The best hyperparameter used in \ref{ex2} and the range $[0.2\bar{\gamma}, 5 \bar{\gamma}]$ is shown in the table.

\begin{table}[!t]
  \caption{Search spaces of hyperparameters used in \ref{ex3}  for DTP.}
  \label{tab:sampling-dtp}
  \begin{center}\begin{small}\begin{sc}
  \begin{tabular}{c|cc}
    \toprule
    Hyperparameter & best $\bar{\gamma}$ & Search space $[0.2\bar{\gamma},   5\bar{\gamma}]$\\
    \midrule
    Learning Rate $\alpha_{f}$ & $4$ & $[0.8, 20]$\\
    Stepsize $\beta$ & $0.002$ & $[0.0004, 0.01]$\\
    Learning Rate $\alpha_{b}$ & $0.004$ & $[0.0008, 0.02]$\\
    \bottomrule
  \end{tabular}
  \end{sc}\end{small}\end{center}
\end{table}

\begin{table}[!t]
  \caption{Search spaces of hyperparameters used in \ref{ex3} for FW-DTP.}
  \label{tab:sampling-fwdtp}
  \begin{center}\begin{small}\begin{sc}
        \begin{tabular}{c|cc}
  \toprule
  Hyperparameter & best $\bar{\gamma}$ & Search space $[0.2\bar{\gamma}, 5\bar{\gamma}]$\\
  \midrule
  Learning Rate $\alpha_{f}$ & $0.02$ & $[0.004, 0.1]$\\
  Stepsize $\beta$ & $0.01$ & $[0.002, 0.05]$\\
  \bottomrule
  \end{tabular}
  \end{sc}\end{small}\end{center}
\end{table}


\end{document}